\documentclass[Afour,sageh,times]{sagej}

\usepackage{amsmath,amssymb,amsfonts}
\usepackage{url,lineno,microtype,subcaption}
\usepackage{graphicx}
\usepackage{textcomp}
\usepackage{arydshln}
\usepackage{xcolor}
\usepackage{natbib}
\usepackage{amsthm}
\theoremstyle{definition}
\newtheorem{definition}{Definition}[section]

\usepackage{algorithm}
\usepackage[noend]{algpseudocode}
\usepackage{etoolbox}
\usepackage{mathtools}

\usepackage{moreverb,url}

\usepackage[colorlinks,bookmarksopen,bookmarksnumbered,citecolor=red,urlcolor=red]{hyperref}

\newcommand\BibTeX{{\rmfamily B\kern-.05em \textsc{i\kern-.025em b}\kern-.08em
T\kern-.1667em\lower.7ex\hbox{E}\kern-.125emX}}

\def\volumeyear{2016}
\hypersetup{draft}

\begin{document}

\runninghead{Smith and Wittkopf}

\title{A Robust Policy Bootstrapping Algorithm for Multi-objective Reinforcement Learning in Non-stationary Environments}

\author{Sherif Abdelfattah, Kathryn Kasmarik, and Jiankun Hu}

\affiliation{School of Engineering and Information Technology,\\
 University of New South Wales, Canberra, ACT, Australia}

\corrauth{Sherif Abdelfattah, School of Engineering and Information Technology,\\
 University of New South Wales, Canberra, ACT, Australia}

\email{sherif.abdelfattah@student.adfa.edu.au}

\begin{abstract}
Multi-objective Markov decision processes are a special kind of multi-objective optimization problem that involves sequential decision making while satisfying the Markov property of stochastic processes. Multi-objective reinforcement learning methods address this kind of problem by fusing the reinforcement learning paradigm with multi-objective optimization techniques. One major drawback of these methods is the lack of adaptability to non-stationary dynamics in the environment. This is because they adopt optimization procedures that assume stationarity in order to evolve a coverage set of policies that can solve the problem. This paper introduces a developmental optimization approach that can evolve the policy coverage set while exploring the preference space over the defined objectives in an online manner. We propose a novel multi-objective reinforcement learning algorithm that can robustly evolve a convex coverage set of policies in an online manner in non-stationary environments. We compare the proposed algorithm with two state-of-the-art multi-objective reinforcement learning algorithms in stationary and non-stationary environments. Results showed that the proposed algorithm significantly outperforms the existing algorithms in non-stationary environments while achieving comparable results in stationary environments.
\end{abstract}


\maketitle

\section{Introduction}
\noindent In reinforcement learning (RL), an agent learns from interactions
with the environment guided by a reward
signal. The objective of the learning process is to find a mapping
from the environment's state space to the action space that maximizes
the expected reward return~\citep{sutton1998reinforcement}. Over the last
decade, RL research has made many breakthroughs such as playing computer
games with human-level performance~\citep{mnih2015human} or beating
human experts in complicated games such as Go~\citep{silver2016mastering}.
This has been achieved by maximizing a single reward function (e.g., the score in games).
However, many real-world sequential decision making problems involve multiple objectives that
may be in conflict with each other. Consider a search and rescue scenario in which an unmanned ground vehicle (UGV) aims at optimizing multiple objectives including minimizing the exposure chance to risks in the environment (i.e., fire or flood), maximizing the rescue ratio of trapped victims, and minimizing the overall search and rescue time. Similarly, a drone in a patrolling scenario trying to maximize battery usability, maximize the patrolled area, and maximize the detection rate of danger/enemy objects. These problems exhibit conflicting objectives that cannot be simultaneously optimized without a compromise. In reinforcement learning and control literature, such problems are known as multi-objective Markov decision processes (MOMDPs). 

Conventional reinforcement learning methods cannot directly tackle the MOMDP problem, as the defined objectives are assumed to be in conflict and can not be simultaneously optimized by a single policy. Rather, multi-objective reinforcement learning (MORL) extends the conventional RL methodology through maximizing a vector of rewards instead of a single reward. Primarily, this is achieved by one of two MORL approaches: the single policy approach; and the multiple policy approach~\citep{roijers2013survey}. 
 
 In the single policy approach, it is assumed that an optimal user's preference can be identified before solving the problem.  Consequently, the multiple objectives can be transformed into a single objective using the supplied user's preference. The main limitations of this approach lie in the difficulty of satisfying its main assumption in many practical multi-objective problems. First, it may be impossible to find an optimal user's preference beforehand. Secondly, it is difficult to deal with changes in the user's preference in real-time, as it is necessary to optimize a different policy after each preference change. Changes in the user's preference can arise if the learning agent needs to deal with different users (such as in computer games or a personal assistant) or due to dealing with changes in the environment's setup (e.g., opponent actions) or in the objective space (e.g., new priorities or new objectives).
 
 Alternatively, the multiple policy approach addresses the MOMDP problem by finding a coverage set of optimal policies that can satisfy any user's preference in solving the problem. This is achieved by performing an intensive policy search procedure that evolves and orders policies based on their performance across the defined objectives. While overcoming the limitations of the single policy approach, this comes with additional costs. First, this approach has higher computational complexity as it requires an intensive interaction with the environment to evolve a set of policies instead of of a single one. Secondly, it assumes stationary environmental dynamics. This makes it inflexible to non-stationary environments, as changes in the environment will demand re-optimization of the evolved policy coverage set.
 
In order to overcome these limitations in the current MORL methods in either dealing with changes in the user's preference, or with the non-stationary environments, we propose a developmental multi-objective optimization method. The main hypothesis behind this method is that, despite the existence of a large set of specialized policies for every possible user's preference, there is possibly a bounded set of steppingstone policies that can bootstrap any specialized policy. In contrast to specialized policies greedily optimized for a specific user's preference and environment, steppingstone policies are dedicated to an interval of user preference. Targeting a coverage set of steppingstone policies that covers the whole user preference space and can be used to bootstrap specialized policies, provides efficient optimization that can work in an online manner and be robust to non-stationary environments as we show experimentally in this paper.

The contribution of this paper is threefold. First, we propose a robust policy bootstrapping (RPB) algorithm that evolves a convex coverage set (CCS) of steppingstone policies in an online manner and utilizes the CCS to bootstrap specialized policies in response to new user preferences or changes in the environmental dynamics. Second, we experimentally evaluate each design decision for the proposed algorithm in order to shed light on the configurable parts that can be changed for different scenarios. Finally, we compare our proposed algorithm with state-of-the-art multi-objective reinforcement learning algorithms in stationary and non-stationary multi-objective environments.

The remainder of this paper is organized as follows. The background section introduces the fundamental concepts and the problem definition. The related work section reviews relevant literature. The methodology section illustrates the proposed algorithm and its workflow. The experimental design section describes the aim, assumptions, and methodology of the experiments. The results section presents the experimental results and discussion. Finally, the conclusion section concludes the work and identifies potential future work.

\section{BACKGROUND}\label{sec:Background}
This section introduces the fundamental concepts necessary for the work and formulates the research problem.

\subsection{Multi-objective Optimization}
The problem of multi-objective optimization includes multiple conflicting objectives that can not be simultaneously optimized without a compromise \citep{Deb2014}. The mathematical formulation of such a problem can be as follows:

\begin{align}
\max\, & (R^{1}\left(\pi\right),R^{2}\left(\pi\right),\,\ldots\,,R^{M}\left(\pi\right))\label{eq:MOO}\\
s.t.\, & g^{j}\left(\pi\right)\leq0,\,j=1,2,\ldots,J\nonumber 
\end{align}

The aim is to optimize the performance of the resultant solutions over a set of reward
functions $\left\{ R^{1}(\pi),R^{2}(\pi),\ldots,R^{M}(\pi)\right\} $, 
given that each function represents an objective $o^{m}\,(m=1,2,\ldots,M)$, the parameter $\pi\in\Pi$ refers to the parameter configuration of the policy (decision variables) to be optimized over the search space $\Pi$, while $\left\{ g^{1}(\pi),g^{2}(\pi),\ldots,g^{J}(\pi)\right\} $ constitutes the set of constraint functions defined in the problem. 

Policies have to be ranked and evaluated based on performance dominance in order to reach the optimal coverage set that can solve the problem while satisfying all possible user's preferences.

\theoremstyle{definition}
\begin{definition}{Dominance:}
\label{def:Dominance}
\textit{``A solution (A) dominates solution (B) if (A) is better than (B) for at least one objective and is equal to (B) for all other objectives.''} \citep{Sherif_18}
\end{definition}

\begin{figure}
\begin{centering}
\includegraphics[scale=0.35]{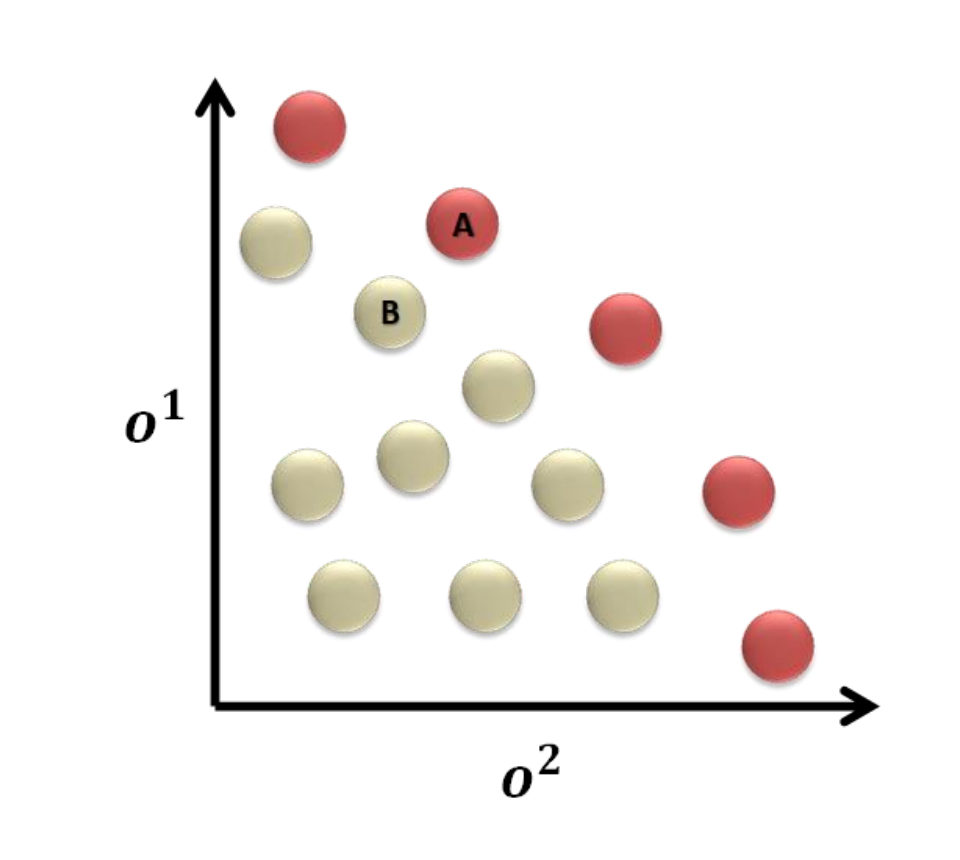}
\par\end{centering}
\caption{The policy search space of a bi-objective problem. The set of non-dominated polices is represented by the red circles, which is known as the Pareto front.}
\label{fig:ParetoFront_Eg}
\end{figure}

Additional illustration of Definition \ref{def:Dominance} is presented in Figure \ref{fig:ParetoFront_Eg}, which shows a two-objective policy search space. It can be noticed that policy (B) is dominated by policy (A). The set of red circles represents the set of non-dominated policies known to as the Pareto front.

The user selects a policy from the set of non-dominated policies based on his/her preference over the defined objectives.

\theoremstyle{definition}
\begin{definition}{Preference:}
\label{def:Preference}
\textit{``A preference is defined as a weight vector with each weight element dedicated to a specific objective $\vec{w}=\left[w^{1},w^{2},\,\ldots\,,w^{M}\right]^{T}$, such that the sum of all the elements equals one $\sum_{m=1}^{M}w^{m}=1$.''} \citep{Sherif_18}
\end{definition}

Given a user's preference, which constitutes a tradeoff among the defined objectives, a scalarization function can be used to formulate a combined objective to be optimized.

\theoremstyle{definition}
\begin{definition}{Scalarization Function:}
\label{def:ScalFunc}
\textit{``A scalarization function $h$, transforms a vector of multiple objectives' values into a single objective scalar value given a preference as parameter $o_{\vec{w}}=h(\vec{o},\vec{w})$.''} \citep{Sherif_18}
\end{definition}

In case of using of linear or piecewise linear scalarization functions \citep{eichfelder2008adaptive}, the non-dominated policies front is referred to as the convex hull (CH).

\theoremstyle{definition}
\begin{definition}{Convex Hull:}
\label{def:CH}
\textit{``A convex hull is a subset of the policy space ($\Pi$) that contains optimal policies that can match any user's preference.''} \citep{Sherif_18}
\begin{center}
$CH(\Pi)=\left\{ \pi:\pi\in\Pi\wedge\exists\vec{w}\,\forall\left(\pi^{\prime}\in\Pi\right)\vec{w}\cdot\vec{r}^{\pi}\geq\vec{w}\cdot\vec{r}^{\pi^{\prime}}\right\} $
\par\end{center}
\end{definition}

\begin{figure*}[tp]
   \begin{minipage}[b]{.5\linewidth}
     \centering
     \includegraphics[width=8cm, height=6cm]{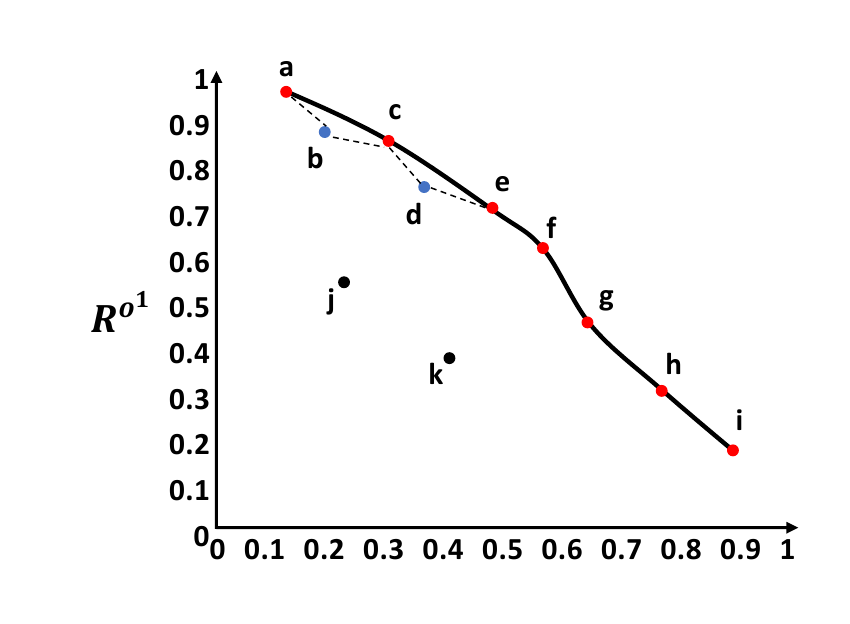}
     \subcaption{}\label{fig:CH_front}
   \end{minipage}
   \hfill
   \begin{minipage}[b]{.5\linewidth}
     \centering
     \includegraphics[width=8cm, height=6cm]{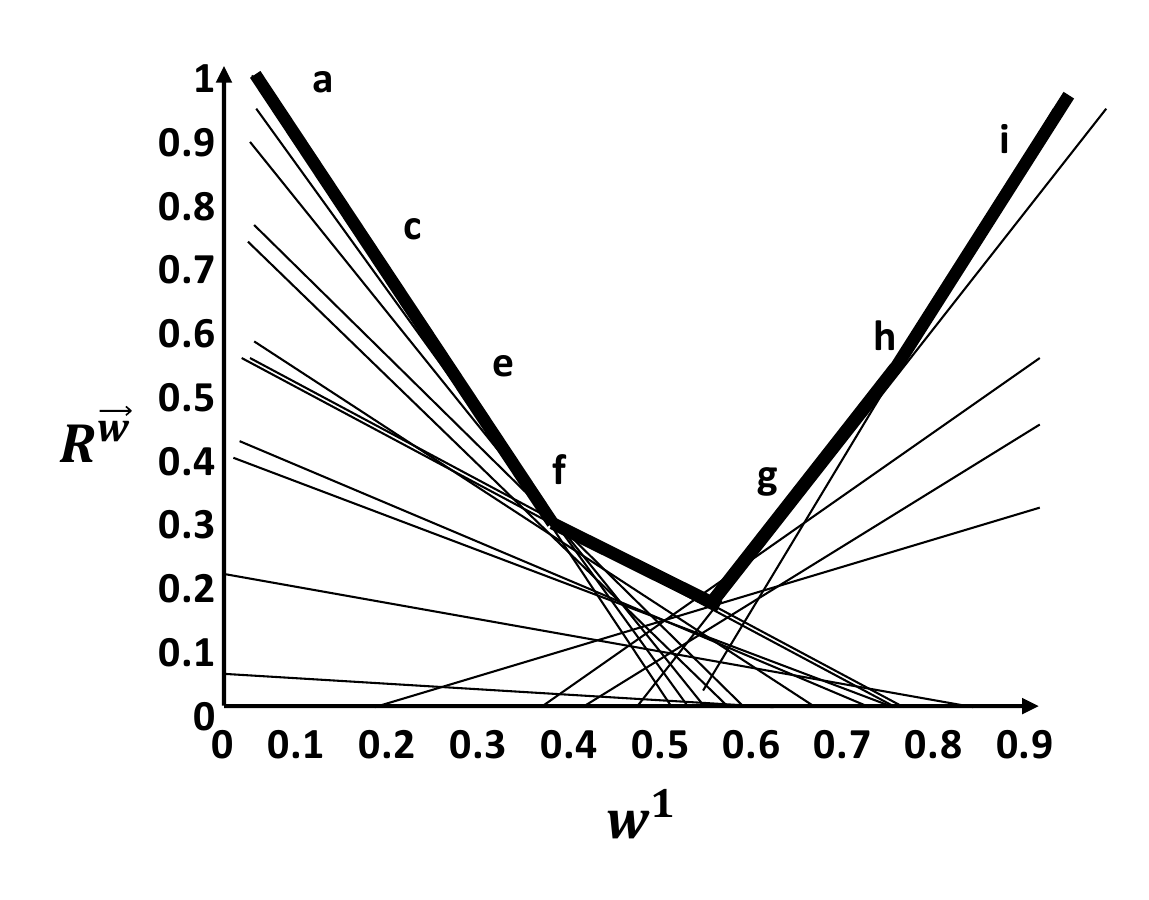}
     \subcaption{}\label{fig:CH_Weights}
   \end{minipage}
   \caption{Comparing the Pareto front surface with the convex hull surface in a bi-objective scenario.\citep{Sherif_18}.}
   \label{fig:CH_Concept}
\end{figure*}

For further illustration of the CH concept, Figure \ref{fig:CH_Concept} visualizes the CH surface using linear scalarization functions over a two objective problem. The axes represent the normalized reward functions for the two objectives. The surface shaped by solid lines, which includes the set of non-dominated policies (i.e., red circles), represents the CH. The non-convex surface shaped by solid and dashed line segments represents the Pareto front which contains all the red and blue points. The non-dominated policy set falling within the CH is depicted as red circles. The blue circles refer to the set of non-dominated policies that falls within the Pareto front and outside the CH. Dominated policies are represented by black circles. 

Figure \ref{fig:CH_Weights} depicts the scalarized reward front produced using a linear scalarization function. Where each line represents a unique preference. The first weight component ($w^{1}$) is shown on the x-axis and the second weight component can be computed from the first component value given the summation to one constraint ($w^{2}=1-w^{1}$). The y-axis shows the scalarized reward value. In Figure \ref{fig:CH_Weights}, the surface of the bold black line segments, which forms a linear piecewise  functions, represents the CH in this scenario.  

As the CH surface can contain redundant policies \citep{roijers2013survey}, we can define a subset of it that contains the minimal number of unique policies that solve the problem, to be the convex coverage set (CCS). 

\theoremstyle{definition}
\begin{definition}{Convex Coverage Set:}
\label{def:CCS}
\textit{``A convex coverage set (CCS) is a subset of the CH that can provide for each preference ($\vec{w}$) a policy whose scalarized reward value is maximal.''} \citep{Sherif_18}
\begin{center}
$CCS\left(\Pi\right)\subseteq CH\left(\Pi\right)\wedge\left(\forall\vec{w}\right)\left(\exists\pi\right)\left(\pi\in CCS\left(\Pi\right)\wedge\forall\left(\pi^{\prime}\in\Pi\right)\vec{w}\cdot\vec{r}^{\pi}\geq\vec{w}\cdot\vec{r}^{\pi^{\prime}}\right)$
\par\end{center}
\end{definition}

\subsection{Multi-objective Markov Decision Processes}

\begin{figure*}[tp]
   \begin{minipage}[]{.5\linewidth}
     \centering
     \includegraphics[scale=0.45]{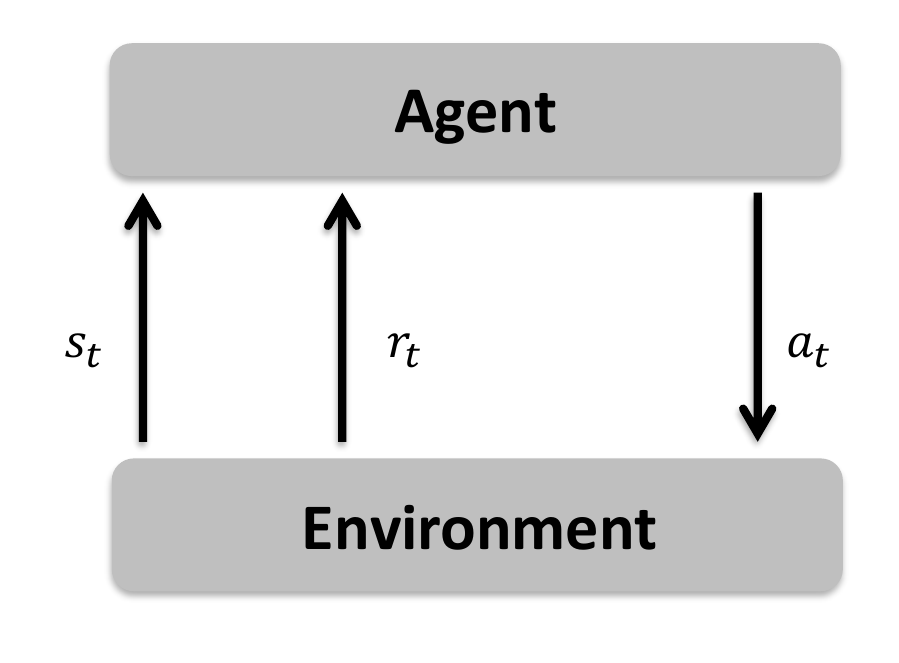}
     \subcaption{}\label{fig:RL}
   \end{minipage}
   \hfill
   \begin{minipage}[]{.5\linewidth}
     \centering
     \includegraphics[scale=0.6]{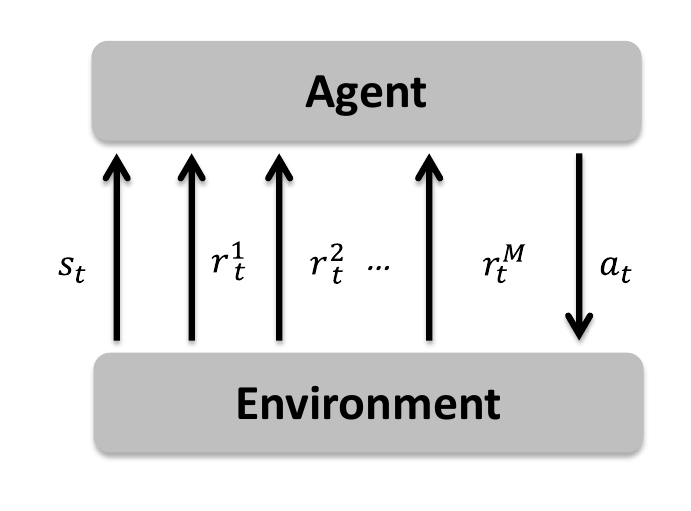}
     \subcaption{}\label{fig:MORL}
   \end{minipage}
   \caption{Comparing single objective Markov decision process (MDP) with the multi-objective variant known as multi-objective Markov decision process (MOMDP). (a) Markov decision process (MDP). (b) Multi-objective Markov decision process (MOMDP) \citep{Sherif_18}.}
   \label{fig:RLMORL}
\end{figure*}

A Markov decision process (MDP) is a sequential planning problem in which the learning agent senses its current state in the environment ($s_t$) at time $t$, performs an action ($a_t$) which results in transition to a new state ($s_{t+1}$) and receives a reward/penalty ($r_{t+1}$) for reaching this new state \citep{Tsitsiklis87}. This paradigm is extended in multi-objective Markov decision processes by having multiple reward signals instead of a single one after performing an action \citep{roijers2013survey}. Figure \ref{fig:RLMORL} shows a comparison between these two problems. The tuple $\left\langle S,A,\mathbb{P_{\mathrm{ss^{'}}}},\vec{R},\mu,\gamma\right\rangle $ represents a MOMDP problem, given that $S$ refers to the state space, the action space is $A$, $\mathbb{P_{\mathrm{ss^{'}}}\mathrm{=Pr(\mathit{s_{t+1}}=\mathit{s^{'}}|\mathit{s_{t}}=\mathit{s,a_{t}}=\mathit{a})}}$ represents the state transition distribution which may have a time varying parameters (dynamics) in non-stationary environments, $\vec{R}\in\mathfrak{\mathbb{R^{\mathrm{M}}}}\,\forall R\,:\,S\times A\times S^{\prime}\rightarrow r\,\in\mathbb{R}$ represents the rewards vector corresponding to   $M$ objectives, $\mathrm{\mu=Pr(s_{0})}$ is the initial state distribution, and finally, $\gamma\in[0,1)$ represents the discounting factor that balances the priority of immediate versus future rewards during the learning process. 

Accordingly, the learning agent aims at maximization the expected return of the scalarized reward signal. Initializing at time $t$ and provided a user's preference $\vec{w}$, this scalarized return can be calculated as: 

\begin{equation}
R_{t}^{\vec{w}}=\sum_{l=0}^{T}\gamma^{l}h\,(\vec{r}_{t+l+1},\vec{w})
\label{Eq:ScalarizedReturn}
\end{equation} 

\noindent such that $T$ refers to the \textit{time horizon} which approaches $\infty$ in the \textit{infinite time horizon} situation. 

\setcounter{subfigure}{0}

\subsection{Problem Definition}
The research problem in this paper can be formulated as: provided a MOMDP problem setup $\left\langle S,A,\mathbb{P_{\mathrm{ss^{'}}}},\vec{R},\mu,\gamma\right\rangle $, we aim at finding the CCS under non-stationary dynamics of the environment (i.e., time varying parameters of state transition distribution) that satisfies the minimal cardinality while maximizing the scalarized reward return for any preference set given at $T$ \textit{time horizon}:
\begin{align}
\begin{split}
\max\;R_{t}^{\vec{w}^{i}} & =\sum_{l=0}^{T}\gamma^{l}h\,(\vec{r}_{t+l+1},\vec{w}^{i})\\
\end{split}
\\
\begin{split}
\min\;\left|CCS\right|\nonumber\\ 
\end{split}
\\
\begin{split}
 & s.t.\:\vec{w}^{i}\in W\,\forall\vec{w}^{i}\in\mathbb{R^{\mathit{M}}\mathit{,\sum_{m=1}^{M}w^{m}}=\mathit{1}}\nonumber 
\end{split}
\end{align}
where $W$ is the set of all legitimate user's preferences over the defined objectives.

\section{RELATED WORK}\label{sec:RelatedWork}

In this section, we review related work in the MORL literature in order to highlight the contribution of this paper.

There are two main methods in MORL literature: single policy methods; and multiple policy methods \citep{roijers2013survey,liu2015multiobjective}. Given a user's preference defined prior to solving the problem, then a single policy can be learned using a scalarized reward function utilizing any single objective reinforcement learning algorithms. In contrast, multiple policy methods search and prioritize the policy search space in order to reach the set of non-dominated policies that can solve the problem given any possible user's preference. We are going to review each of these methods as follows.

\subsubsection{Single Policy Methods}

Many single method techniques utilize scalarization functions in order to combine the multiple defined objectives into a single objective. Kasimbeyli et al. \citep{Kasimbeyli2015} discussed the properties of different scalarization functions in solving multi-objective optimization problems. Moffaert et al. \citep{Moffaert2013} introduced a variant of the Q-learning algorithm \citep{Watkins1992} that utilizes the Chebyshev function for reward scalarization solving an MOMDP grid-world scenario. Castelletti et al. \citep{castelletti2013} used  non-linear scalarization functions with a random exploration of the weight space for optimizing the workflow of irrigation resource management systems. Lizotte et al. \citep{lizotte2010efficient} introduced linear scalarization with an algorithm that adopts the classical value iteration algorithm in order to rank actions for finite state space problems. Perny and Weng \citep{Perny_2010} utilized a linear programming method that incorporates Chebyshev scalarization to address a MOMDP problem. 

Alternatively, other methods utilize linear programming techniques that follow lexicographic ordering of objectives \citep{MARCHI1992355}. Ogryczak, Perny, and Weng \citep{Ogryczak_2011} further developed the aforementioned method by adopting a regret technique instead of reward scalarization for action ranking. The regret value was derived for each objective given an anchor point and the ranking is done based on the summed regret from all the defined objectives. \citep{Feinberg_95,Altman_1999} proposed an alternative approach based on constrained multi-objective optimization which transforms the MOMDP problem into a MDP problem by focusing the optimization process on one objective while dealing with the remaining objectives as constraints.

\subsubsection{Multiple Policy Methods}

Techniques of user's preference elicitation learn the user's preference gradually by sampling from different policy trajectories and observing the user's feedback in order to adjust future action selection \citep{Mousseau2015}. One method of this group was proposed by Akrour et al. \citep{Akrour2011} where the preference of domain experts was acquired within the policy optimization using an algorithm named preference-based policy learning (PPL). The algorithm assumed a defined parameterization of the policy that can be sampled to draw different trajectories. The preference of the domain expert is inferred implicitly by asking them to rank the performed trajectories, which is utilized to guide the next sampling iteration of trajectories. A similar approach was introduced by F\"urnkranz et al. \citep{Furnkranz2012} which assumes that the Pareto front of optimal policies is found before questioning the expert's preference. 

 Another group of methods adopts evolutionary optimization techniques \citep{Abraham:2005} for searching in the policy space to tackle the MOMDP problem. An evolutionary algorithm was proposed by Busa-Fekete et al. \citep{Busa-Fekete2014} for finding the Pareto front of non-dominated policies. Afterwards, the learning agent will perform the action recommended by one of the policies in the Pareto front based on the user's feedback in ranking different sampled trajectories. 
 
 Other methods utilize systematic preference exploration techniques to evolve a coverage set of policies that can solve the MOMDP problem. These methods achieved comparable results to the evolutionary optimization alternatives with more efficient computation complexity \citep{roijers2013survey}. Roijers, Whiteson, and Oliehoek \citep{Roijers2014} introduced an algorithm named Optimistic Linear Support (OLS) which evolves an approximate coverage set by systematically investigating various user's preferences over the defined objectives. As an example, given a two-objectives scenario, the algorithm explores the two edge preferences (i.e., $[0.1,0.9]$, $[0.9,0.1]$) and generates two corresponding policies using a reinforcement learning algorithm (i.e., Q-learning). The performance of the evolved policies is contrasted using a predefined threshold parameter where the one that exceeds it will be added to the final solution set. The algorithm will continue this procedure by further dividing the two edge preferences using a median point and repeating the same selection technique until no more policies exceeding the threshold parameter can be found. 
 
Policy lexicographic ordering has been explored by G{\'a}bor, Kalm{\'a}r, and Szepesv{\'a}ri \citep{gabor1998multi} in their Threshold Lexicographic Ordering (TLO) algorithm which ranks policies based on a predefined threshold value. For each sampled policy, the TLO algorithm performs the action that achieves the maximum reward value exceeding another predefined threshold in any of the reward functions or the maximum reward value if all actions are below the threshold.

OLS and TLO algorithms have been widely adopted in many relevant literature \citep{roijers2015point,geibel2006reinforcement,mossalam2016multi} to evolve convex coverage sets of non-dominated policies. Therefore, they form a good representation for the current state-of-the-art methods and can clearly contrast the performance of our proposed algorithm.

 In a previous work \citep{Sherif_18}, we embarked on investigating the non-stationary MOMDP environments by proposing a new benchmark environment which poses non-stationary state transition dynamics and proposing a multi-objective reinforcement learning algorithm based on the fuzzy segmentation of the preference space. In this work, we propose a generic and simpler algorithm and investigate the different options and design decisions not explored in the previous work. We utilize our previously introduced non-stationary benchmark in the evaluation of the proposed algorithm.

 \section{Robust Policy Bootstrapping For Solving MOMDPs}\label{sec:Methodology}

The proposed robust policy bootstrapping (RPB) algorithm aims at maximizing the learning agent's robustness
to perturbations in the user's preferences by evolving a coverage
set of robust policies that can cope with these perturbations.

\begin{figure}
\begin{centering}
\includegraphics[scale=0.4]{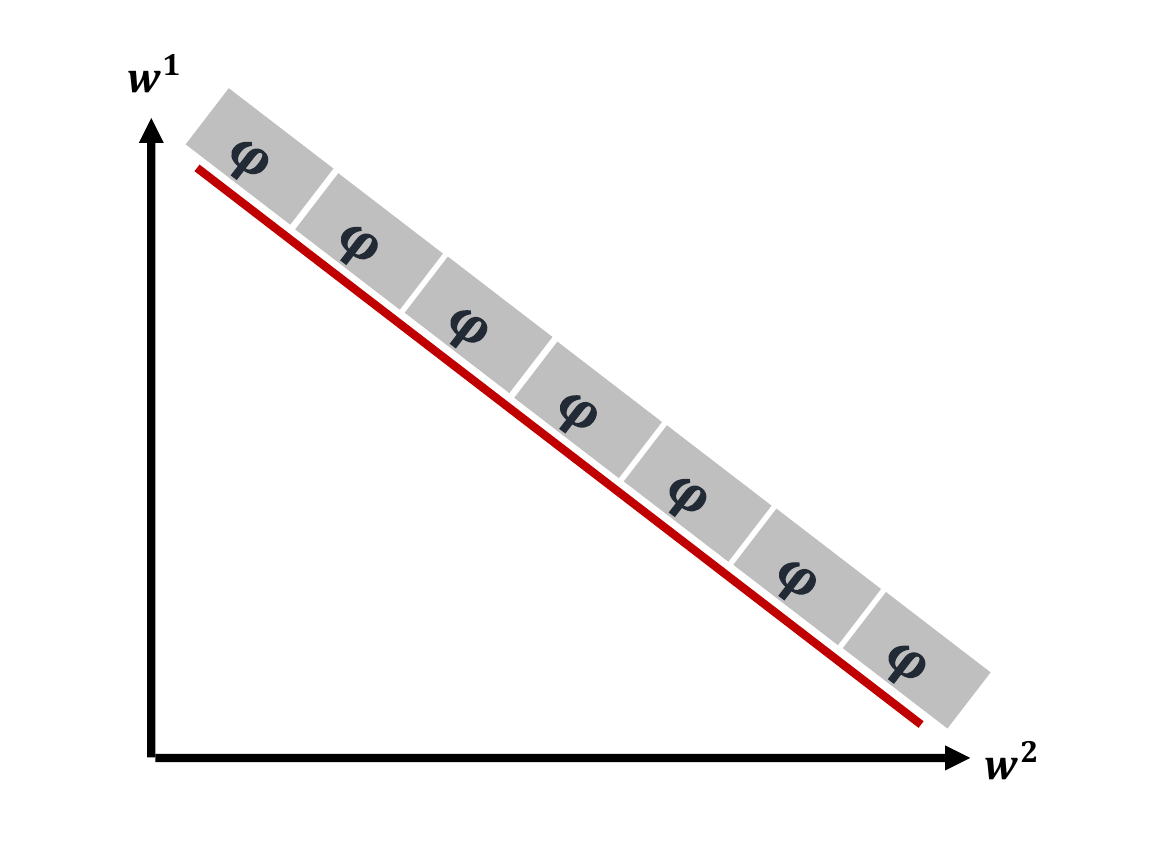}
\par\end{centering}
\caption{Dividing the linear scalarization of two objectives into regions using the threshold value ($\varphi$).}
\label{fig:RPB_Threshold}
\end{figure}

As mentioned previously, our main assumption is that while there may be a large number
of policies, each corresponding to a specific user's preference,
a smaller and finite number of robust policies can offer the best
steppingstone for any change in the preference. To find this coverage set
of robust policies, we assume a preference threshold value ($\varphi$)
that divides the linear scalarization of the defined reward functions into a number $G$ of regions (see Figure \ref{fig:RPB_Threshold}). Then,
for each region, a single steppingstone policy is evolved that maximizes a robustness metric, which measures the stability of its behaviour.

This paper first describes the generic design of our methodology. Then, We experimentally evaluate a number of design options to  shed light on configurable parts that can be changed in different applications.
 
\begin{figure*}
\noindent \begin{centering}
\includegraphics[scale=0.8]{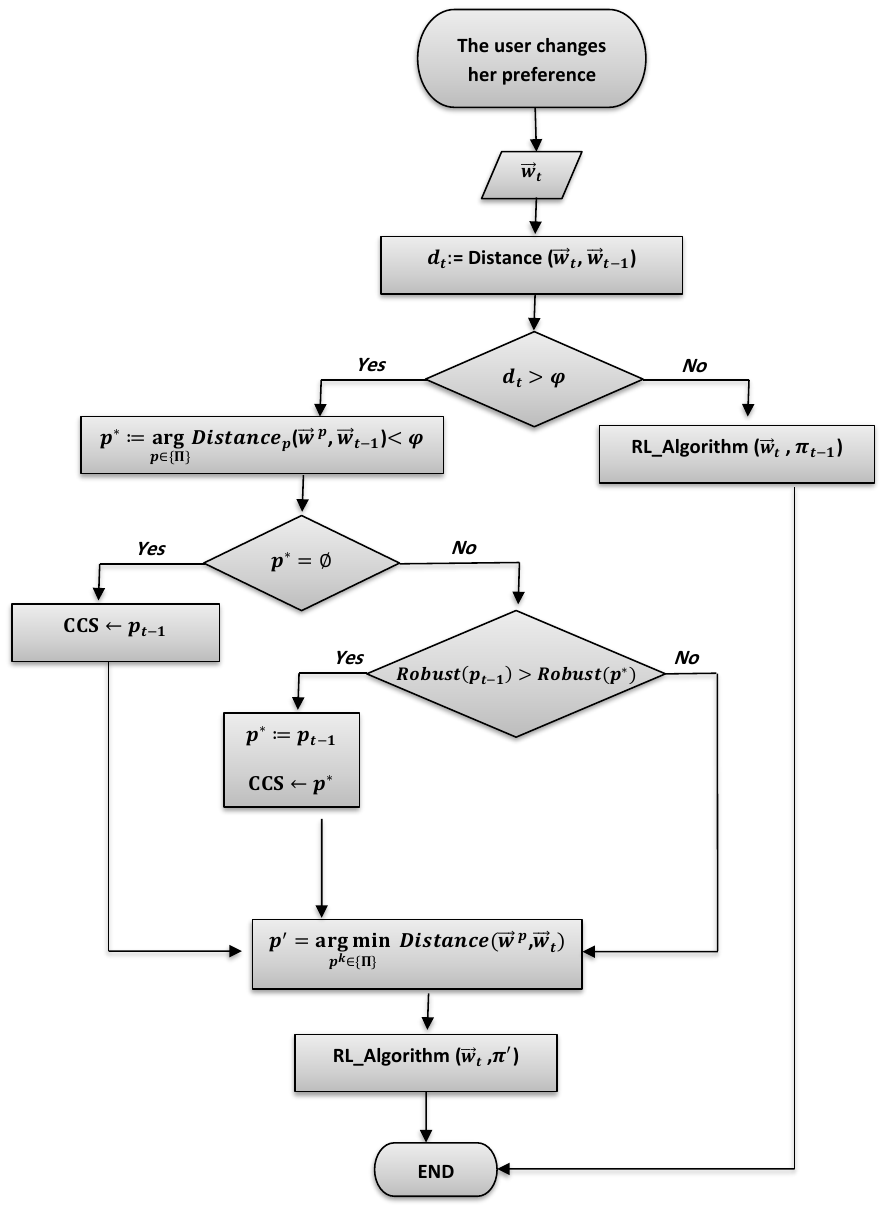}
\par\end{centering}
\caption{A flowchart diagram describing the generic RPB algorithm workflow (Algorithm \ref{alg:RPB}).}
\label{fig:flowchart_generic}

\end{figure*}

 Figure \ref{fig:flowchart_generic} shows the generic flowchart of our proposed algorithm. The flow starts after a change occur in the user's preference represented by a weight vector over the defined reward functions. Then, the distance between the previous working preference ($\overrightarrow{w}_{t-1}$) and the new one ($\overrightarrow{w}_{t}$) is measured using a vector distance function. If the distance is less than the significance threshold ($\varphi$), (which means that we are still in the same preference region) then the current policy is used to respond to the new preference and policy optimization using the RL algorithm will continue. Otherwise, the policy bootstrapping mechanism is activated. The first step in this mechanism is to store the steppingstone policy for the previous preference region in the $CCS$. To achieve that we search in the existing $CCS$ for a steppingstone policy dedicated to the previous preference region. If a steppingstone policy is found, then it is compared with the current policy on the basis of robustness value. Finally, the best one is stored in the $CCS$. Alternatively, in the case that there is no steppingstone policy found for the previous preference region, then the current policy is directly assigned to it and saved in the $CCS$. For each steppingstone policy
$p^{k}$, we store three parameters $\left\langle \pi^{k},\vec{w}^{k},\beta^{k}\right\rangle $.
Where $\pi^{k}$
is the adopted reinforcement learning algorithm's parameters (e.g., Q-values in Q-learning algorithm), $\vec{w}^{k}$
is the preference corresponding to this policy, and $\beta^{k}$ is the
robustness metric value. Finally, we search in the $CCS$ for the steppingstone policy with the minimum distance to the new preference. This policy is used to bootstrap the ensuing policy optimization procedure using the reinforcement learning algorithm.

 As we are dealing with grid world benchmark environments, we adopt a scalarized version of Q-learning~\citep{Sherif_18, Watkins1992}
for solving the MOMDP task using scalarization given the weights vector
as depicted in Algorithm~\ref{alg:S-QL}. Q-learning has been successfully deployed to optimize policies for such environments effectively \citep{Q1,Q2,Q3,Q4,Q5}. Accordingly, the policy parameters $\pi^{k}$ stored in the $CCS$ are the $Q(s,a)$ values for each state and action pairs.  We refer to this algorithm as SQ-L. 

It has to be noted that the proposed RPB algorithm is not limited to the use of Q-learning only as any reinforcement learning algorithm \citep{szepesvari2010} can be used instead of it without affecting the workflow. In other words, changing the reinforcement learning algorithm will only change the policy parameterization $\pi$. For example, changing Q-learning with a policy gradient reinforcement learning algorithm \citep{sutton2000policy} will change $\pi$ from $Q(s,a)$ values to weight matrices of a neural network. 

\begin{algorithm}
\caption{Scalarized Q-Learning (SQ-L) \citep{Sherif_18}}\label{alg:S-QL}

\begin{algorithmic}[1]
  \Require A preference $\vec{w},\pi^{init}$
  \If{$\pi^{init}=\phi$ }
  \State Initialize $Q(s,a)\:\forall\,s\in S,\:a\in A(s)$ arbitrarily
  \Else
  \State Initialize $Q(s,a)\:\forall\,s\in S,\:a\in A(s)$ from $\pi^{init}$
  \EndIf
  \Repeat
  \For{each episode} 
  \State Initialize $S$
  \For{each time point $t$}
  \State Take $a_{t}$ from $s_{t}$ using policy derived from $Q$ (e.g., $\epsilon$-greedy) , observe $\vec{r}_{t+1},\:s_{t+1}$ 
  \State Calculate scalarized reward $\rho=\vec{w}\cdot\vec{r}_{t+1}$ 
  \State $Q(s_{t},a_{t})\leftarrow Q(s_{t},a_{t})+\alpha\left[\rho+\gamma max_{a^{\prime}}Q(s_{t+1},a^{\prime})-Q(s_{t},a_{t})\right]$ 
  \State $s\leftarrow s_{t+1}$ 
  \EndFor
  \EndFor
  \Until{$s$ is terminal}
\end{algorithmic}
\end{algorithm}

The pseudocode of the proposed robust policy bootstrapping algorithm is presented in Algorithm \ref{alg:RPB}.  

\begin{algorithm}
\caption{Robust Policy Bootstrapping (RPB)}\label{alg:RPB}
\begin{algorithmic}[1]
  \Require Preferences at times $t$ and $t-1$ ($\mathbf{\protect\overrightarrow{w}}_{t},\mathbf{\protect\overrightarrow{w}}_{t-1}$).
  \State $d_{t}\coloneqq d\left(\mathbf{\overrightarrow{w}}_{t},\mathbf{\overrightarrow{w}}_{t-1}\right)$ 
  \If{$d_{t}>\varphi$}
  \State set $p^{*}$ equal to the policy with preference distance $d\left(\mathrm{\mathbf{\overrightarrow{w}^{*}},\mathbf{\overrightarrow{w}}_{t-1}}\right)\leq\varphi$ from $CCS$ 
  \If{$p^{*}=\emptyset$}
  \State Store $p_{t-1}$ in $CCS$ 
  \ElsIf{$\beta(p_{t-1})>\beta(p^{*})$}
  \State Replace $p^{*}$ with $p_{t-1}$ in $CCS$ 
  \EndIf
  \State Retrieve the best policy $p^{\prime}=\underset{p^{k}\in CCS}{\arg\min}\left[d\left(\mathbf{\overrightarrow{w}}_{t},\mathbf{\overrightarrow{w}}^{k}\right)\right]$ 
  \EndIf
  \State Follow the Scalarized Q-Learning algorithm, SQ-L($\overrightarrow{w}_{t},\,\pi^{\prime}$)
     
\end{algorithmic}
\end{algorithm}

 There are three main design decisions that need to be configured in the previously described generic procedure as follows. 
  
 \begin{enumerate}
\item The distance function $d\left(\mathbf{\overrightarrow{w}}_{t},\mathbf{\overrightarrow{w}}_{t-1}\right)$ that measures the difference between user preferences.
\item The metric $\beta(p)$ for measuring the robustness of generated policies.
\item The value for the preference significance threshold parameter $\varphi$.
\end{enumerate}

We are going to conduct a separate empirical analysis for each of these design decision in order to highlight the impact of different configurations.

\section{EXPERIMENTAL DESIGN}\label{sec:ExperimentalDesign}

In this section, we present our experimental design for assessing
the performance of the proposed algorithm. 

\subsection{Environments}

\begin{figure*}[tp]
   \begin{minipage}[]{.32\linewidth}
     \centering
     \includegraphics[width=4.2cm,height=4.2cm]{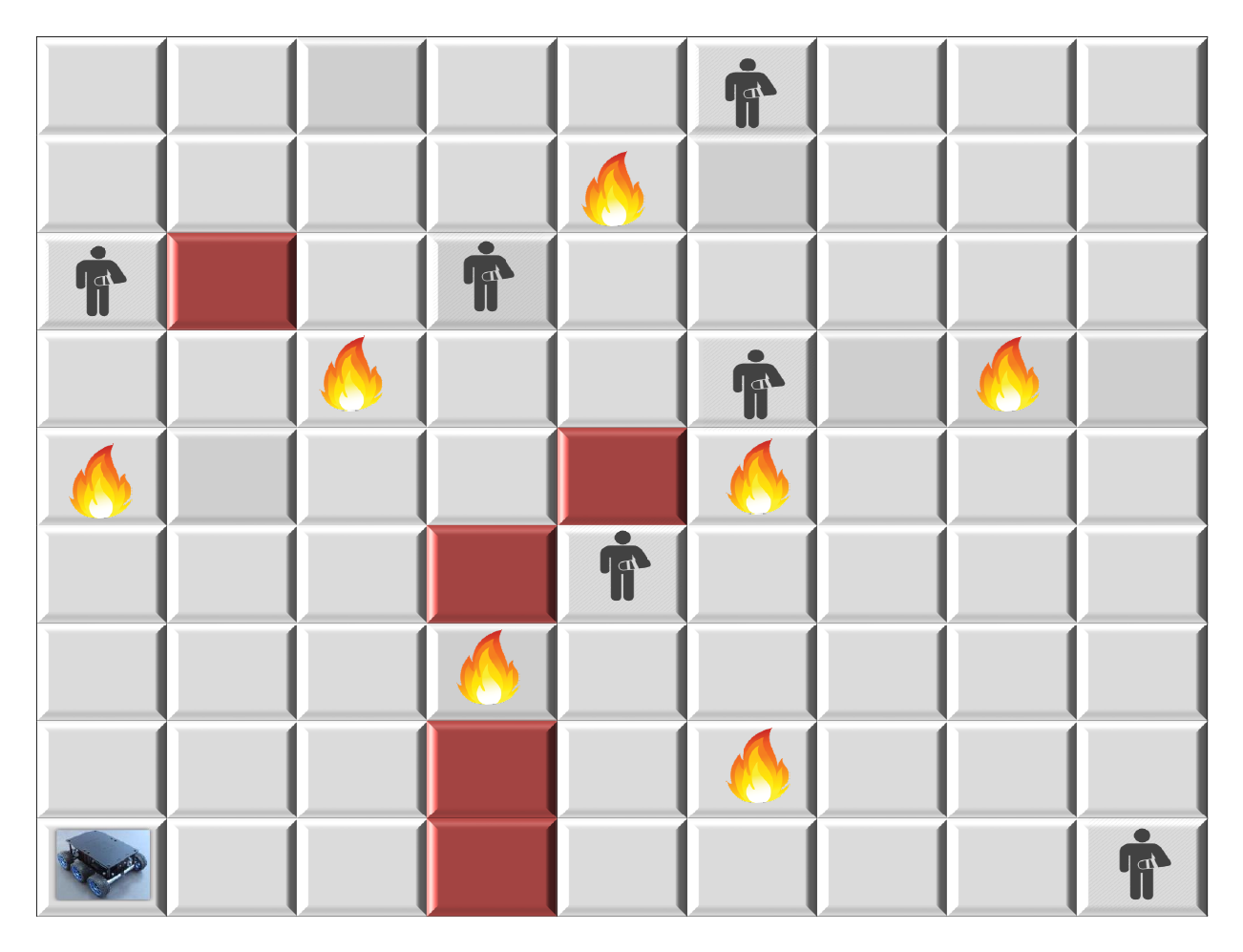}
     \subcaption{}\label{fig:SAR}
   \end{minipage}
   \begin{minipage}[]{.32\linewidth}
     \centering
     \includegraphics[width=4cm,height=4cm]{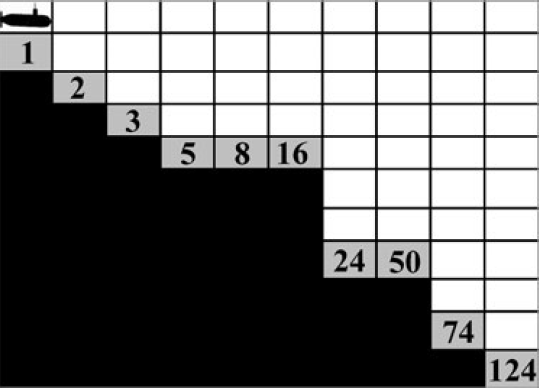}
     \subcaption{}\label{fig:DST}
   \end{minipage}
   \begin{minipage}[]{.32\linewidth}
     \centering
     \includegraphics[width=4cm,height=4cm]{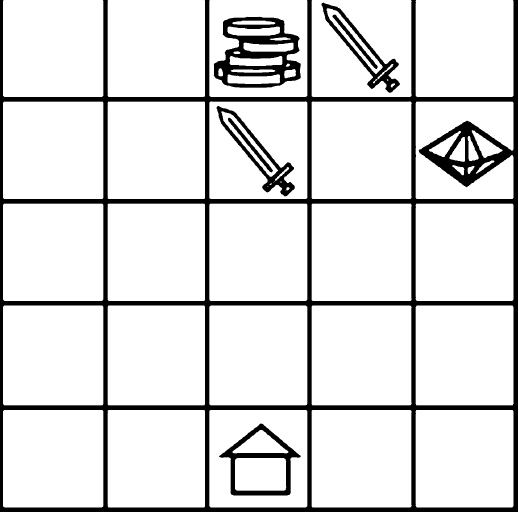}
     \subcaption{}\label{fig:RG}
   \end{minipage}
   \caption{Graphical representation of the benchmark environments. (a) The search and rescue (SAR) environment. (b) The deep sea treasure (DST) environment. (c) The resource gathering (RG) environment \citep{Sherif_18}.}
   \label{fig:Envs}
\end{figure*}

In this work, a benchmark of three MOMDP environments is utilized. This benchmark includes a resource gathering environment, deep-sea treasure environment, and a search and
rescue environment. The former environments are well-established in relevant literature \citep{Vamplew2011}, while the latter one was proposed in our previous work \citep{Sherif_18}. Latter environment has stochastic
state transition dynamics necessary for examining performance in non-stationary environments. A graphical representation of these environments is shown in Figure \ref{fig:Envs}.

\setcounter{subfigure}{0}

\hfill

\subsubsection{Search and Rescue (SAR) Environment}

\hfill 

\textbf{State Space: }SAR environment is a grid-world with dimensions $9\times9$ that simulates a task in which the learning agent has to search for trapped victims while avoiding exposure to fire. Accordingly, the state representation is defined as $\left\langle X,Y,F,O,H\right\rangle $, where $X,\,Y$ indicate the latest position features, and $F,\,O,\,H$ are boolean flags for the existence of fire, an obstacle, or victim in the current position. As a stochastic feature of the state transition in this environment, each victim has a random death time $\xi_{i},\,i\in\left\{ 1,2,3,\ldots,N\right\} $
for $N$ victims. The episode ends when all victims are dead or rescued.

\textbf{Action Space: }The action space is $A=\left\{ MoveEast,\,MoveWest,\,MoveNorth,\,MoveSouth\right\} $
with one square per movement. Attempting to move into a location occupied by an obstacle fails (i.e., the agent remains at its current location and incurs a penalty).

\textbf{Reward Functions: } This environment includes two main objectives: minimizing the time to finish the task, and minimizing destruction by fire. Accordingly, there are two corresponding reward functions $\ensuremath{\vec{r}=\left[r^{\mathit{fire}},\,\mathit{r}^{\mathit{time}}\right],\,\vec{r}\in\mathbb{R\mathrm{^{\mathit{2}}}}}$. Every time the agent collides with a fire it gets a penalty of $-5$ and $0$ elsewhere, while there is a constant time penalty for each step of $-1$.

\hfill 

\subsubsection{Deep Sea Treasure (DST) Environment}

\hfill

\textbf{State Space: } This grid-world environment has dimensions of $10\times11$. It simulates an undersea treasure search scenario in which a submarine agent is trying to maximize the found treasure value. The state representation is defined by the features of the current location $\left\langle X,Y\right\rangle $. The episode ends when a treasure is found. 

\textbf{Action Space: } The learning agent can move in one of the four directions at each step  $A=\left\{ Left,\,Right,\,Up,\,Down\right\} $ where moving in directions that make the agent go outside the gird will not change the latest state.

\textbf{Reward Functions: } The submarine has two conflicting objectives of maximizing the resource reward while minimizing task's time. The reward vector for these objectives is $\vec{r}=\left[\,\mathit{r}^{\mathit{time}},\,\mathit{r}^{\mathit{treasure}}\right],\,\vec{r}\in\mathbb{R\mathrm{^{\mathit{2}}}}$. The value of $r^{time}$ is a constant equal to $-1$ on every action, while treasure reward $r^{treasure}$ will be unique for each treasure. 

\hfill 

\subsubsection{Resources Gathering (RG) Environment}

\hfill

\textbf{State Space: } The dimensions of this grid-world environment are $5\times5$. The learning agent aims at acquiring resources and returning back home without being attacked by an enemy. The representation of the state space is $\left\langle X,Y,G,Y,E\right\rangle $, provided that $X,\,Y$ are the features of the current position, and $G,Y,E$ are boolean flags referring to the existence of gold, gem, or enemy in the current position. This environment has stochastic state transition with probability of enemy attack of $10\%$. After enemy attacks, the agent will lose any gathered resources and will be returned back to its home. The episode ends when the agent safely returns to the home location or when an attack occurs.

\textbf{Action Space: } The learning agent has the ability to move to one of the four directions $A=\left\{ MoveEast,\,MoveWest,\,MoveNorth,\,MoveSouth\right\} $ by one step at a time.

\textbf{Reward Functions: } The learning agent has two objectives in this environment which are to maximize the reward value of collected resources and to minimize exposure to attacks from the enemy. These are represented as $\vec{r}=\left[\,\mathit{r}^{\mathit{resources}},\,\mathit{r}^{enemy}\right],\,\vec{r}\in\mathbb{R\mathrm{^{\mathit{2}}}}$, where $r^{resources}$ equals $1$ for each gathered resource, and every enemy attack will result in a penalty $r^{enemy}$ of $-1$.

\subsection{Comparison Algorithms}

 We compare our proposed algorithm with three MORL algorithms: 1) Optimistic Linear
Support (OLS) \citep{Roijers2014}, 2) Threshold Lexicographic Ordering
(TLO) \citep{gabor1998multi} and 3) Robust Fuzzy Policy Bootstrapping (RFPB) \citep{Sherif_18}. In addition, we add the (SQ-L) algorithm (see Algorithm \ref{alg:S-QL}) to the comparison as a baseline using a random policy initialization after each preference change. We explored an alternative policy initialization approach for the SQ-L algorithm that adopts the parameters of policy optimized for the last preference. We found that given a uniform preference sampling, there was no significant difference between these two policy initialization approaches.

As both the OLS and TLO algorithms require an offline simulation phase to evolve their coverage sets, we run them until convergence before comparing with our algorithm in the execution phase. For the parameter configuration of the OLS, TLO, and RFPB algorithms we utilize the same configuration in \citep{Roijers2014}, \citep{geibel2006reinforcement}, and \citep{Sherif_18} respectively. For the proposed RPB algorithm we conduct empirical analysis in order to compare the effect of different configurations.

\subsection{Experiments}

To shed light on the design decisions and to assess the performance of our RPB algorithm, we conduct five experiments. These experiments include empirical analysis for the three main design decisions: 1) preference significance threshold, 2) robustness metric and 3) and distance function; and two additional experiments for contrasting the performance of the proposed algorithm with state-of-the-art methods in MORL literature in stationary and non-stationary environments.

\hfill

\subsubsection{Empirical Analysis for the Preference Significance Threshold ($\varphi$)}
\label{sec:thresholdVals}
\hfill

\textbf{Aim: } This experiment aims at assessing the optimal value of the preference significance threshold ($\varphi$) in each experimental environment.

\textbf{Method: } We evaluate 10 different threshold values that gradually increase from $0.05$ to $0.5$. For each run, we execute the 10 uniformly sampled preferences in Table \ref{tbl:U} giving each 800 episodes. We run 15 independent runs for each environment. 

\textbf{Evaluation Criteria: } We compare the reward loss values after switching the preference for each threshold value over the 15 independent runs. The one that achieves the lowest value on this metric is the best fit for the environment.

\hfill

\bgroup
\def\arraystretch{1.5}
\begin{table*}[tp]
\caption{The set of uniformly sampled user preferences utilized in the experimental
design.}
\begin{centering}
\begin{tabular}{c c c c c c c c c c}
\hline 
{Preference} & $P_{1}$ & $P_{2}$ & $P_{3}$ & $P_{4}$ & $P_{5}$ & $P_{6}$ & $P_{7}$ & $P_{8}$ & $P_{9}$\tabularnewline
\hline  
$w_{1}$ & 0.66 & 0.33 & 0.28 & 0.54 & 0.68 & 0.44 & 0.88 & 0.65 & 0.48\tabularnewline 
$w_{2}$ & 0.34 & 0.67 & 0.72 & 0.46 & 0.32 & 0.56 & 0.12 & 0.35 & 0.52\tabularnewline
\hline 
\end{tabular}
\par\end{centering}

\label{tbl:U}
\end{table*}
\egroup

\subsubsection{Empirical Analysis for the Robustness Metric ($\beta$)}

\hfill

\textbf{Aim: } The aim of this experiment is to assess the robustness metric to use in the design of the RPB algorithm.

\textbf{Method: } For investigating the different variations of policy robustness metrics, we evaluated five candidate metrics for each environment. We selected these metrics based relevant literature for policy robustness evaluation \citep{bui2012robustness,Chow_NIPS2015,Covariance_2014,deb2006introducing,Ehrgott201417,jen2003stable}. The metrics are presented in Table \ref{tbl:Robust_Metrics}. The first metric measures the stability of the policy's behaviour in terms of mean of rewards divided by standard deviation. The second metric measures the dispersion of a probability distribution (i.e., the observed reward values in our case), it is the ratio of variance to the mean  and it is called index of dispersion (IoD). The third metric measures the degree of variability with respect to the mean of the population. It is calculated as the ratio between the standard deviation and the mean and it is referred to as The coefficient of variation (CV). The fourth metric is the entropy of the reward distribution. It is derived from the concepts of Information theory. Finally, the fifth metric is the regret, defined as the difference between the reward mean of the current policy and the reward mean of the optimum policy. While this metric has the potential to guide the policy search given a ground truth (optimum policy) it is not applicable in many scenarios in which an optimum policy cannot be known prior solving the problem. For the last metric evaluation, we utilize the policies generated by the OLS algorithm as optimum policies to compare with. We separate the regret metric with a dotted line in Table \ref{tbl:Robust_Metrics} in order to indicate the difference in its assumption compared to the rest of the metrics.

\bgroup
\def\arraystretch{1.5}
\begin{table*}[tp]
\caption{Robustness metrics utilized in the empirical analysis for ($\beta$) design decision}
\begin{centering}
\begin{tabular}{c c}
\hline 
Metric & Equation \tabularnewline
\hline 
Stability & $\frac{\mu}{\sigma}$ \tabularnewline 
Index of dispersion (IoD) & $\frac{\sigma^{2}}{\mu}$ \tabularnewline
Coefficient of variation (CV) & $\frac{\sigma}{\mu}$ \tabularnewline
Entropy & $-\sum_{i=1}^{n}P(x_{i})\log_{b}P(x_{i})$ \tabularnewline
\hdashline 
Regret & $\mu_{\pi^{*}}-\mu_{\pi}$ \tabularnewline
\hline 
\end{tabular}
\par\end{centering}

\label{tbl:Robust_Metrics}
\end{table*}
\egroup

\textbf{Evaluation Criteria: } We compare the five metrics by running 15 independent runs for each metric. Each run includes 800 episodes for each preference in the $9$ sampled preferences in Table \ref{tbl:U}. Then, we calculate the median reward value for each preference and sum the medians to get one representative value for each run. Finally, we average the 15 independent sums and visualize the average with standard deviation bars.

\hfill

\subsubsection{Empirical Analysis for the Distance Function ($d\left(\protect\overrightarrow{w}_{t-1},\protect\overrightarrow{w}_{t}\right)$)}

\hfill

\textbf{Aim: } In this experiment, we evaluate the effectiveness of various distance functions over all experimental environments. 

\textbf{Method: } We evaluate four well-known distance functions including the Euclidean, Hamming, Cosine, and Manhattan distance functions. The equations for these functions are illustrated in Table \ref{tbl:Distance_Funcs}

\bgroup
\def\arraystretch{1.5}
\begin{table*}[tp]
\caption{A list of distance functions evaluated in the empirical analysis}
\begin{centering}
\begin{tabular}{c c}
\hline 
Distance Function & Equation \tabularnewline
\hline 
Euclidean & $\sqrt{\sum_{m=1}^{M}\left(w_{t-1}^{m}-w_{t}^{m}\right)^{2}}$ \tabularnewline 
Hamming & $\sum_{m=1}^{M}\left[w_{t-1}^{m}\neq w_{t}^{m}\right]$ \tabularnewline
Cosine & $\frac{\mathbf{w_{t-1}}\cdot\mathbf{w_{t}}}{\left\Vert \mathbf{w_{t-1}}\right\Vert \left\Vert \mathbf{\mathbf{w_{t}}}\right\Vert }$ \tabularnewline
Manhattan & $\sum_{m=1}^{M}\left|w_{t-1}^{m}-w_{t}^{m}\right|$ \tabularnewline
\hline 
\end{tabular}
\par\end{centering}

\label{tbl:Distance_Funcs}
\end{table*}
\egroup

\textbf{Evaluation Criteria: } We followed a similar evaluation criteria as in the robustness metric empirical analysis. We run 15 independent runs for each metric. Each run includes 800 episodes for each preference in the $9$ sampled preferences in Table \ref{tbl:U}. Then, we calculate the median reward value for each preference and sum the medians to get one representative value for each run. Finally, we average the 15 independent sums and visualize the average with standard deviation bars.

\hfill

\subsubsection{Performance Evaluation in Stationary Environments}

\hfill

\textbf{Aim:} The aim of this experiment is to assess the impact of user's preference changes on each of the comparison methods while neutralizing the impact of the environment's dynamics changes by operating in a stationary environment. In this experiment, the environment's dynamics (i.e., distribution of objects) is fixed during each run. Therefore, only the start location of the agent is randomized at the beginning of each run.  In addition, the two comparative MORL algorithms (OLS and TLO) assume a stationary environment setup, therefore, this experiment guarantees a fair comparison by including the best scenario for them. 

\textbf{Method: } We compare the four algorithms based on a set of ($9$) random preferences sampled uniformly from the two-dimensional weight space for the three experimental environments. Table \ref{tbl:U} presents the set of the randomly sampled preferences. We execute $30$ runs each with a different initial distribution of objects. Each preference is given a total of $800$ episodes and the average reward value over the last $50$ episodes is used to compare the four algorithms. As the OLS and TLO algorithms are working in an offline mode, we firstly, run their training procedures till convergence (evolving the CCS), then, we compare them with the other algorithms. The RPB algorithm is configured based on the outcomes of the empirical analysis of each design configuration.

\textbf{Evaluation Criteria: } We evaluate the performance of each algorithm using two metrics: the average reward value over the last 50 episodes for each preference; and the average reward loss after preference change. The former metric reflects the performance level of the generated policy for each algorithm in response to the user's preference in terms of reward value after convergence. While the latter metric shows the robustness of each algorithm to changes in the user's preference in terms of reward loss.  

\begin{figure}
\begin{centering}
\includegraphics[scale=0.4]{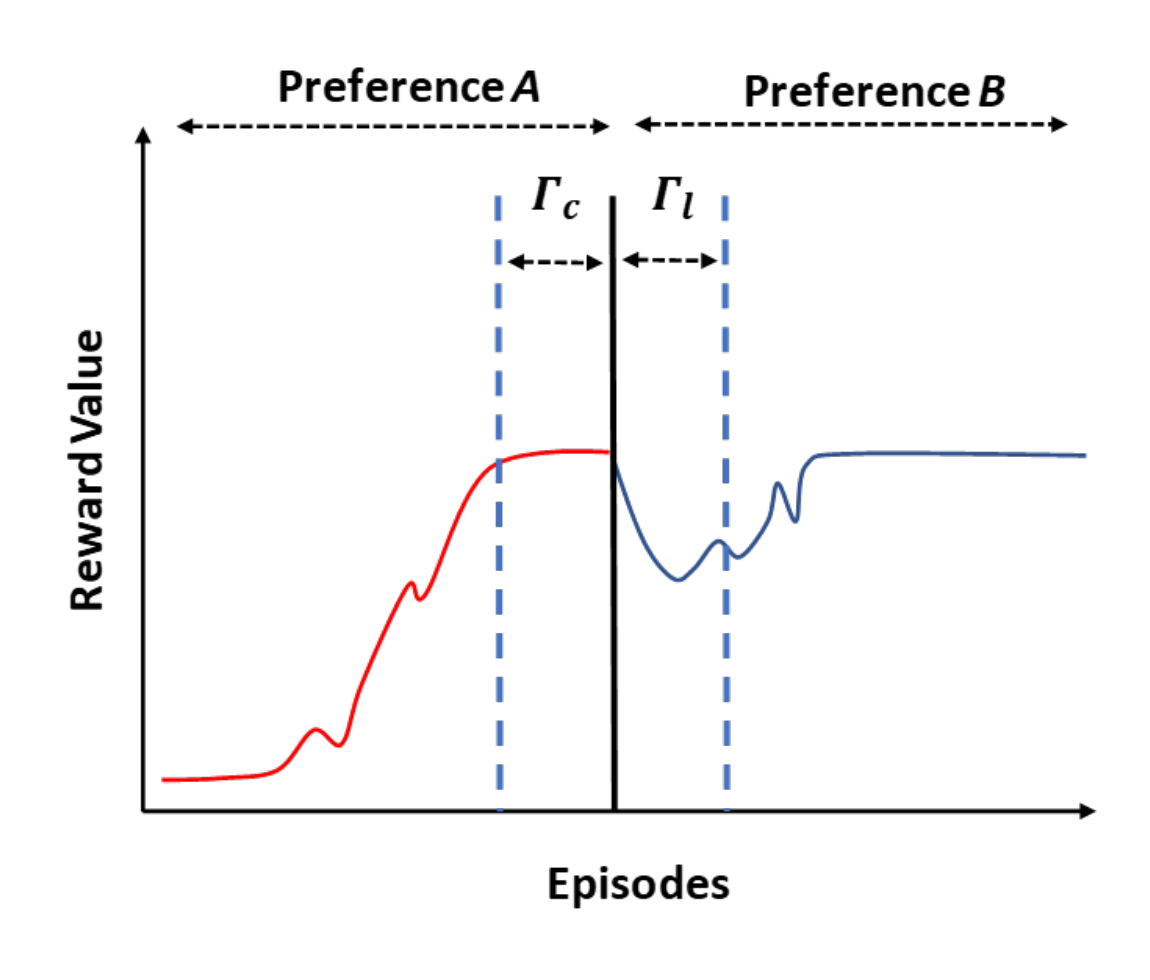}
\par\end{centering}
\caption{A diagram that that illustrates the two evaluation metrics utilized in our experimental design. The average reward value over the last 50 episodes is represented by $\Gamma_{c}$. While the average reward loss after 
change from preference (A) to preference (B) is represented by $\left(\Gamma_{c}-\Gamma_{l}\right)$. }

\label{fig:Diagram}
\end{figure}

For further illustration of these two metrics, Figure \ref{fig:Diagram} shows a diagram for the average reward signal
over a single preference change from preference (A) to preference (B).
The value $\Gamma_{c}$ is the average reward value over the last $50$
episodes of preference (A). While $\Gamma_{l}$ is the average reward value
over the first 50 episodes of preference (B). We utilize $\Gamma_{c}$
to show the average reward value after convergence and $\left(\Gamma_{c}-\Gamma_{l}\right)$
to show the average loss value after the preference change from (A)
to (B).

\hfill

\subsubsection{Performance Evaluation in Non-stationary Environments}

\textbf{Aim:} This experiment aims at evaluating the impact of the environment dynamics change, while using the same preference change pattern, therefore, the performance level of each comparison algorithm in non-stationary environments can be well contrasted. In order to achieve that we allow the environment's setup to change within each run by allowing $25$\% of the objects (e.g., fire, victim, obstacle, treasure) to change their locations randomly every $100$ episodes.

\textbf{Method: } We utilize the same method as in the stationary environment experiment.

\textbf{Evaluation Criteria: } We use the same evaluation metrics as in the stationary environment experiment.

\section{RESULTS AND DISCUSSION}\label{sec:Results}

In this section, we are going to present and discuss the results of each of the five experiments included in the experimental design. 

\subsection{Empirical Analysis for the Preference Significance Threshold ($\varphi$)}

\begin{figure*}[tp]
   \begin{minipage}[]{0.5\linewidth}
     \centering
     \includegraphics[scale=0.35]{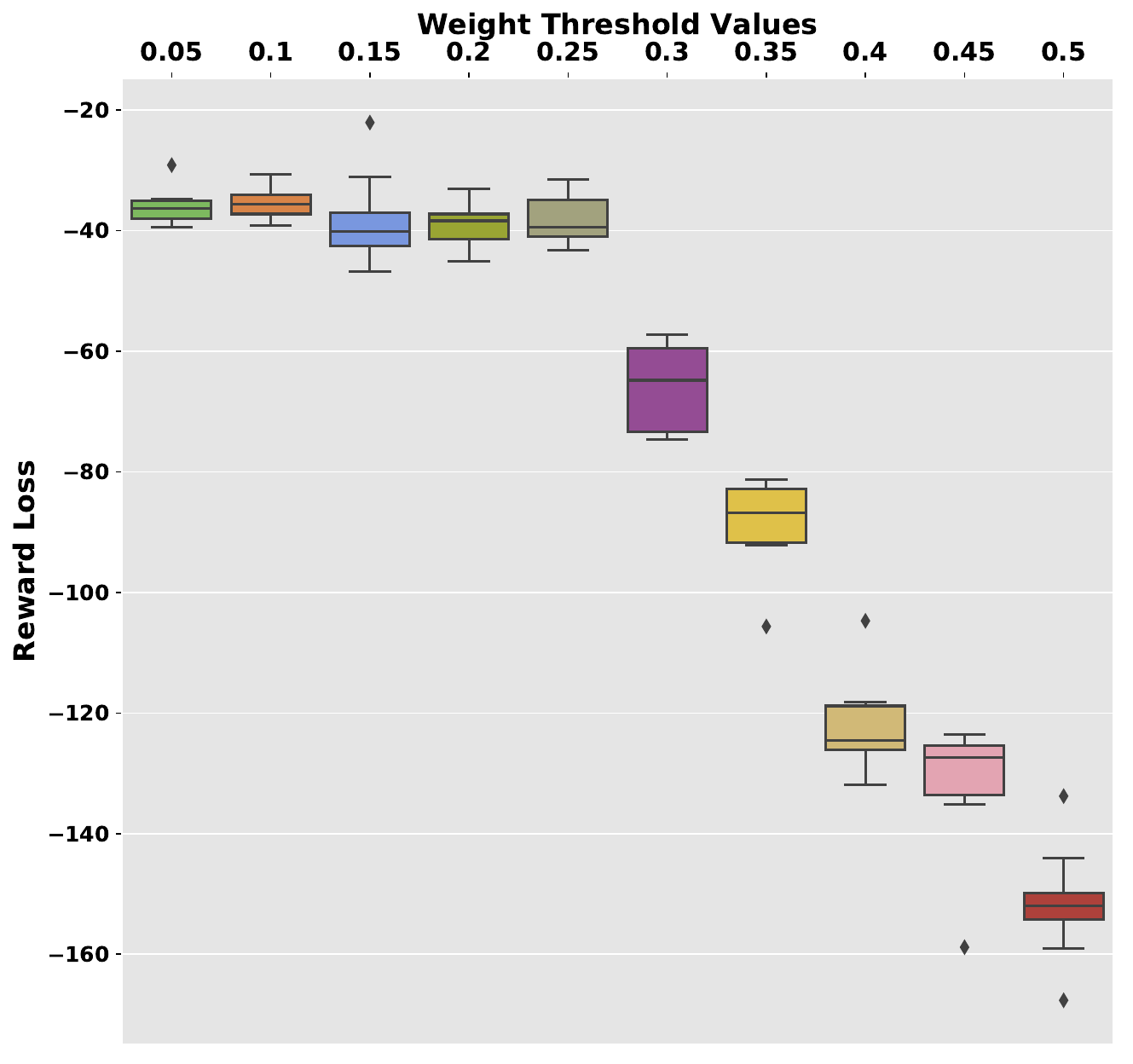}
     \subcaption{}\label{fig:SAR}
   \end{minipage}
   \begin{minipage}[]{0.5\linewidth}
     \centering
     \includegraphics[scale=0.35]{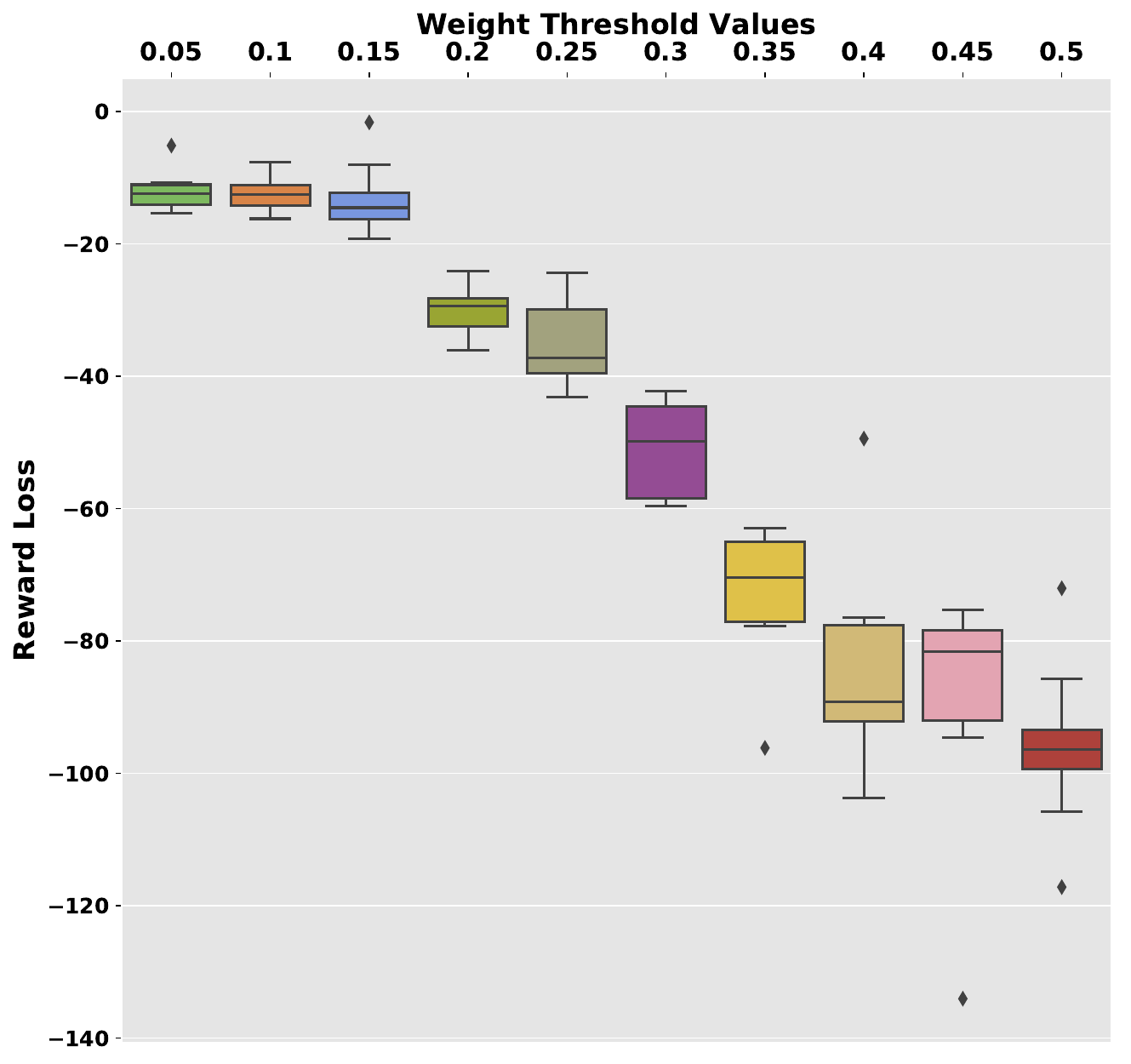}
     \subcaption{}\label{fig:DST}
   \end{minipage}
   \begin{minipage}[]{1\linewidth}
     \centering
     \includegraphics[scale=0.35]{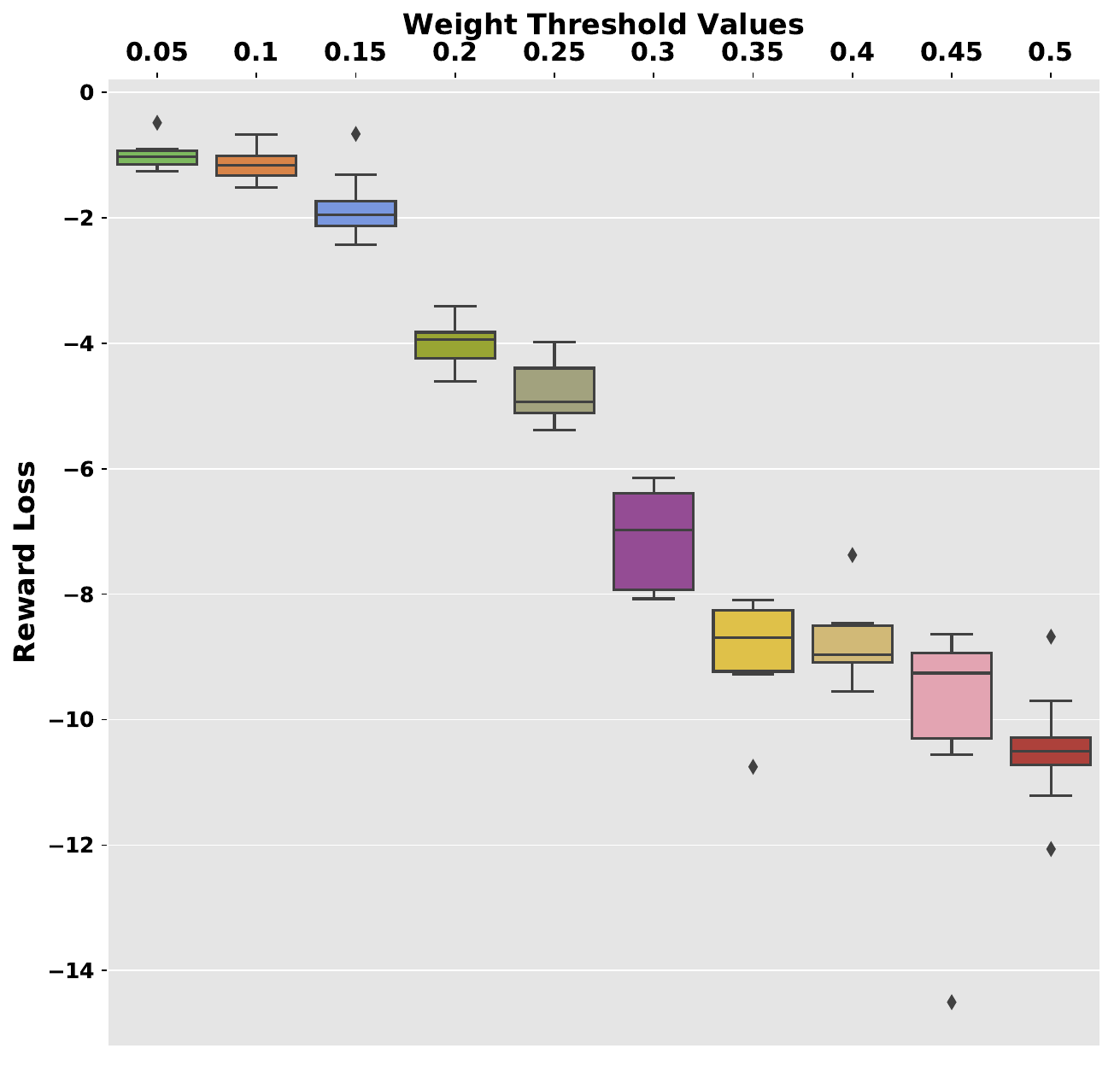}
     \subcaption{}\label{fig:RG}
   \end{minipage}
   \caption{Empirical analysis results for determining the optimal ($\varphi$) threshold value of the RPM algorithm for each experimental environment. The boxplots represent the distribution of reward loss values after switching preferences for each threshold value. (a) Results for the SAR environment. (b) Results for the DST environment. (c) Results for the RG environment.}
   \label{fig:PhiParam}
\end{figure*}

Figure \ref{fig:PhiParam} shows the average reward loss after preference change for the preference distance threshold
parameter ($\varphi$) values for each experimental environment. For the
SAR environment, the value of ($0.25$) was found to be the optimal
as significant reward loss was observed for higher values. While for
the DST and RG environments, the optimum value for ($\varphi$) was
found to be ($0.15$). These results indicate that the there is no one optimum threshold value that can fit all scenarios. In other words, the preference threshold parameter needs to be tuned for each different environment based on its dynamics and the defined reward functions. This finding opens the way for further future enhancements to the proposed algorithm through ways that automatically tune this threshold parameter (our previously proposed algorithm \citep{Sherif_18}, while more complex, is able to do this). The RPB algorithm will utilize the identified $\varphi$ values for each environment during comparison to other MORL algorithms.   

\setcounter{subfigure}{0}

\subsection{Empirical Analysis for the Robustness Metric ($\beta$)}

\begin{figure*}[tp]
   \begin{minipage}[]{0.5\linewidth}
     \centering
     \includegraphics[scale=0.35]{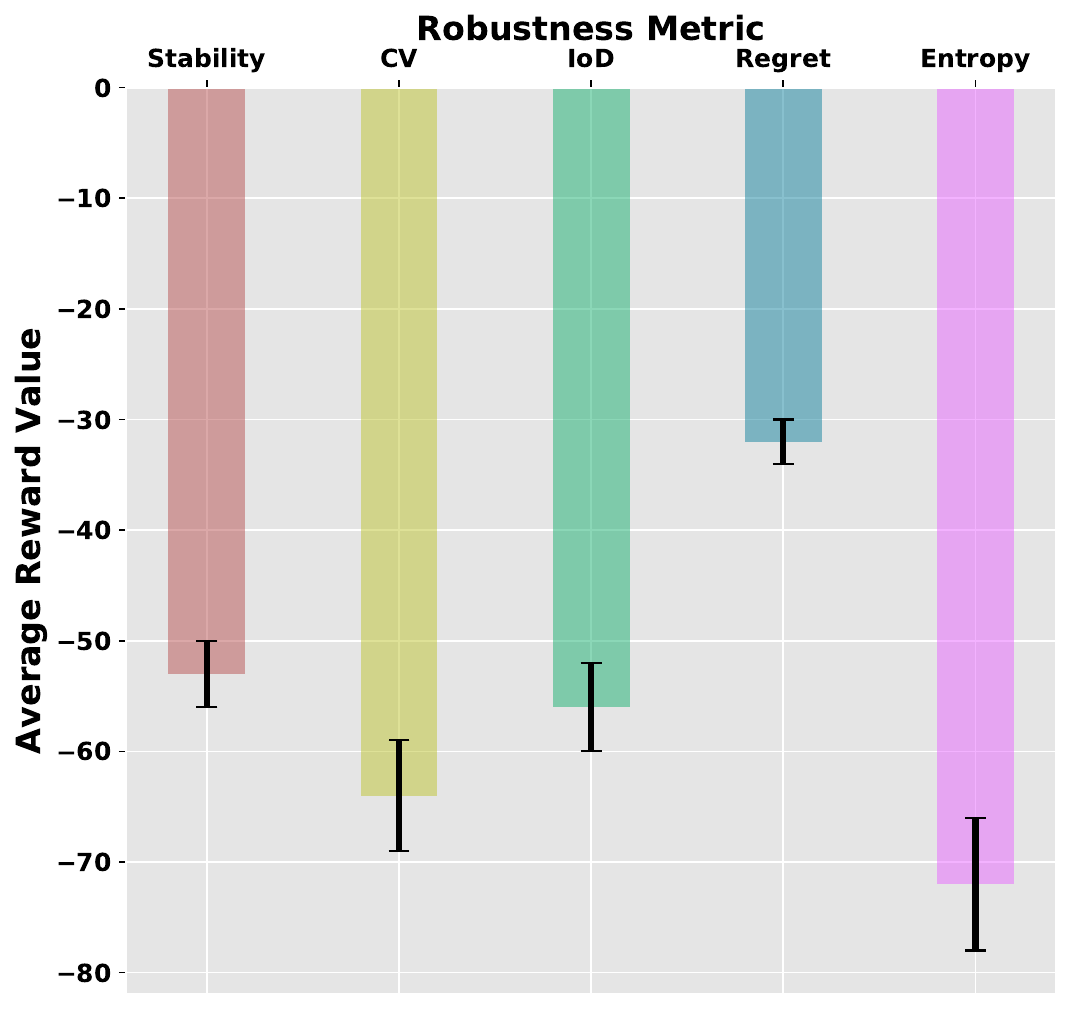}
     \subcaption{}\label{fig:SAR}
   \end{minipage}
   \begin{minipage}[]{0.5\linewidth}
     \centering
     \includegraphics[scale=0.35]{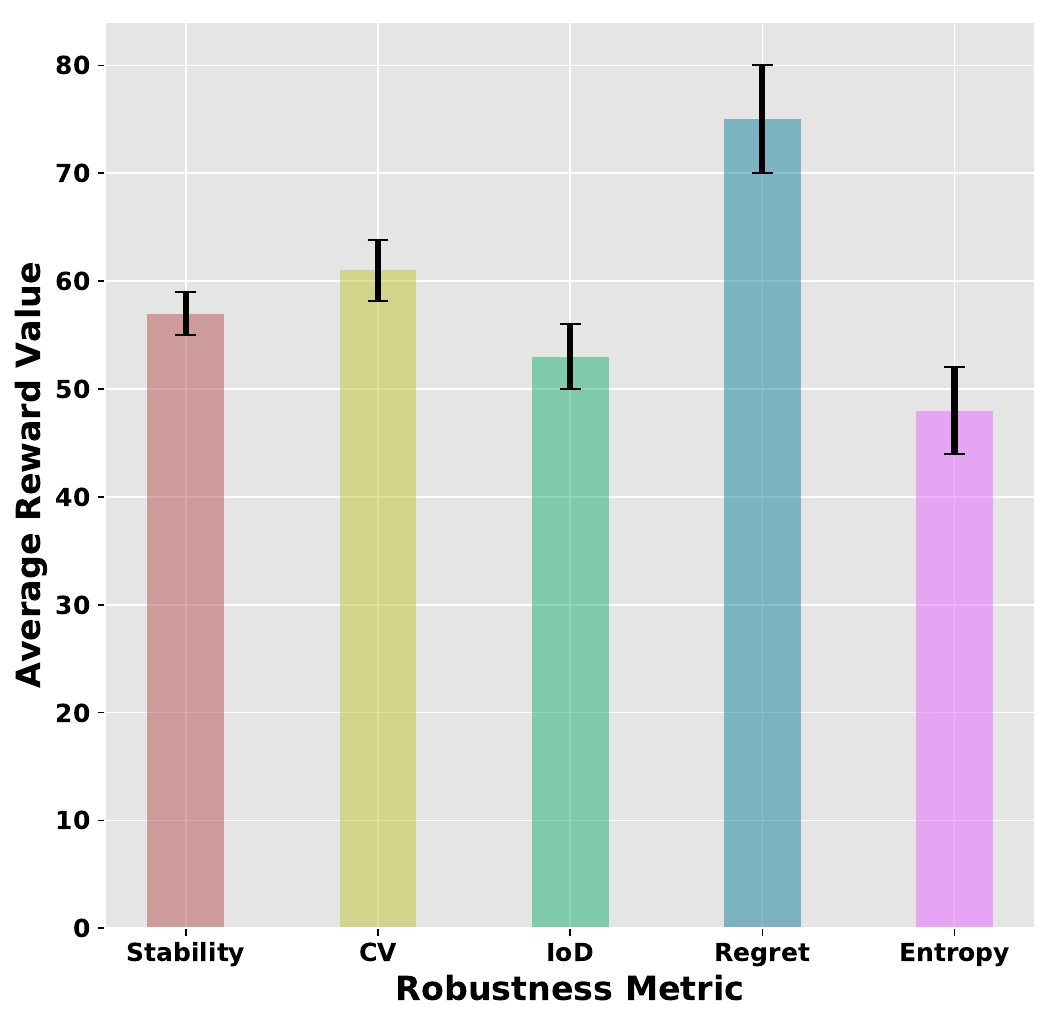}
     \subcaption{}\label{fig:DST}
   \end{minipage}
   \begin{minipage}[]{1\linewidth}
     \centering
     \includegraphics[scale=0.35]{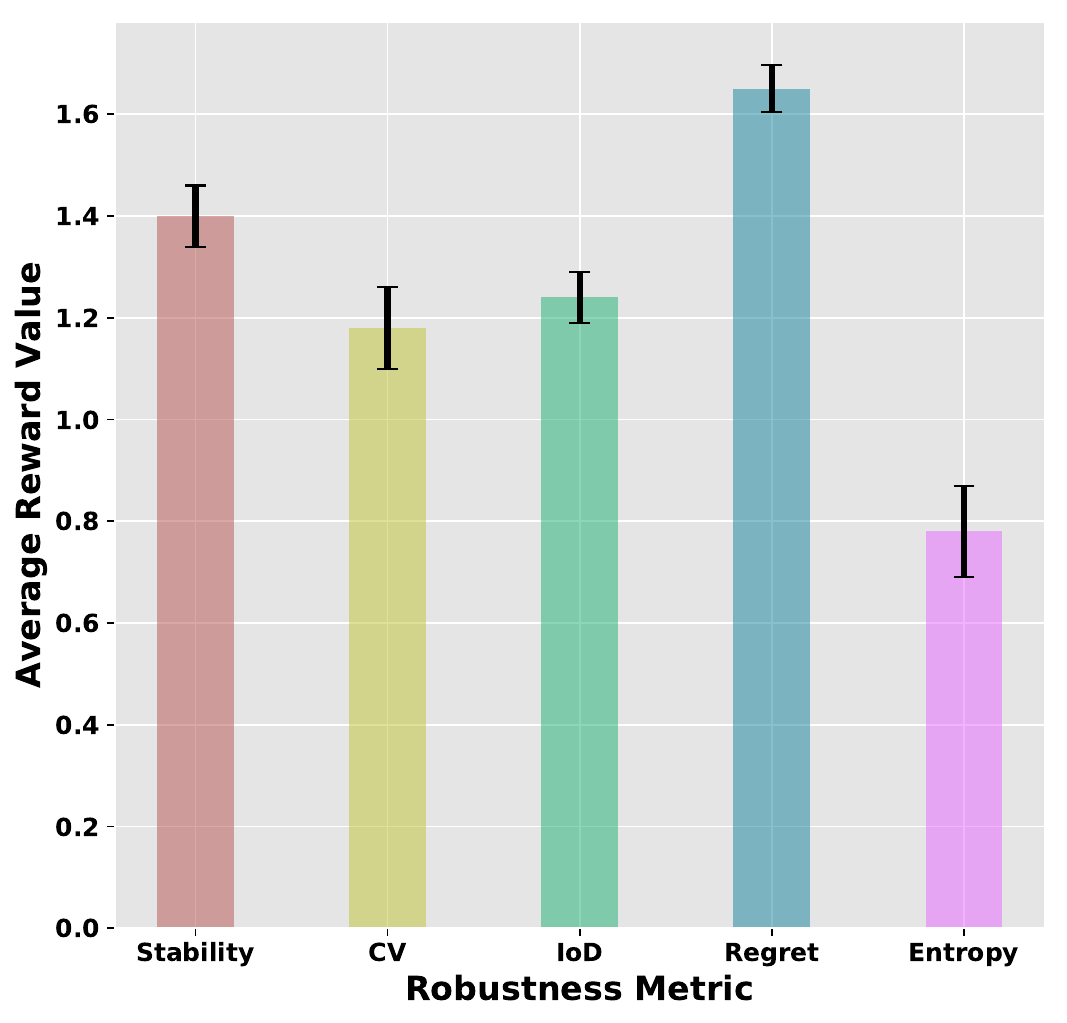}
     \subcaption{}\label{fig:RG}
   \end{minipage}
   \caption{Empirical analysis results for determining the robustness metric to be used in the RPM algorithm for each experimental environment. (a) Results for the SAR environment. (b) Results for the DST environment. (c) Results for the RG environment. The error bars represent the standard deviation.}
   \label{fig:Robust_Metrics}
\end{figure*}

Figure \ref{fig:Robust_Metrics} shows the comparison results for the robustness metrics. While the regret metric achieved the best results in terms of average reward value over the three experimental environment, it is not applicable in many scenarios in which the CCS of policies is not defined beforehand. The stability metric achieved the best overall results in comparison to the other three remaining metrics. Thus, the RPB algorithm will use the stability metric during comparison with other algorithms.  

\setcounter{subfigure}{0}

\subsection{Empirical Analysis for the Distance Function ($d\left(\protect\overrightarrow{w}_{t-1},\protect\overrightarrow{w}_{t}\right)$)}

\begin{figure*}[tp]
   \begin{minipage}[]{0.5\linewidth}
     \centering
     \includegraphics[scale=0.35]{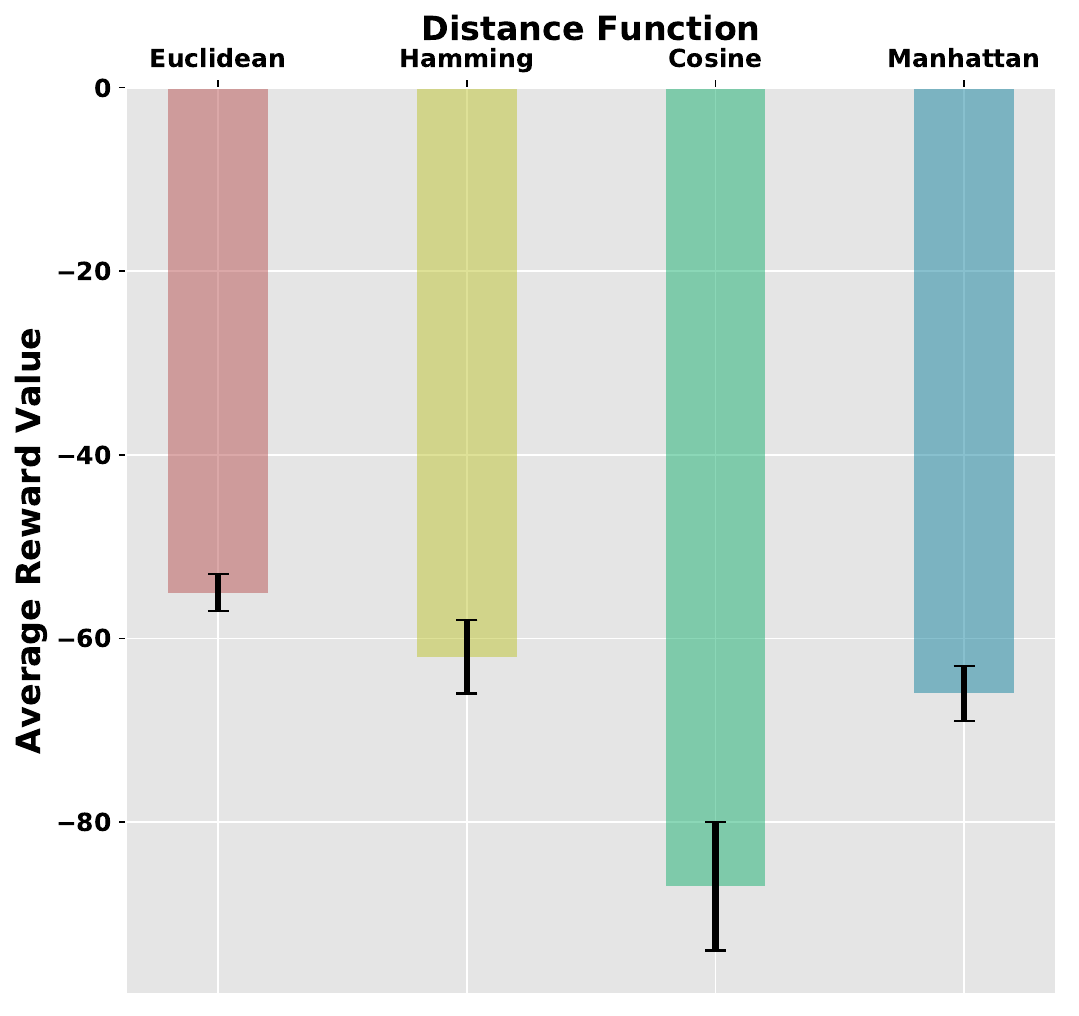}
     \subcaption{}\label{fig:SAR}
   \end{minipage}
   \begin{minipage}[]{0.5\linewidth}
     \centering
     \includegraphics[scale=0.35]{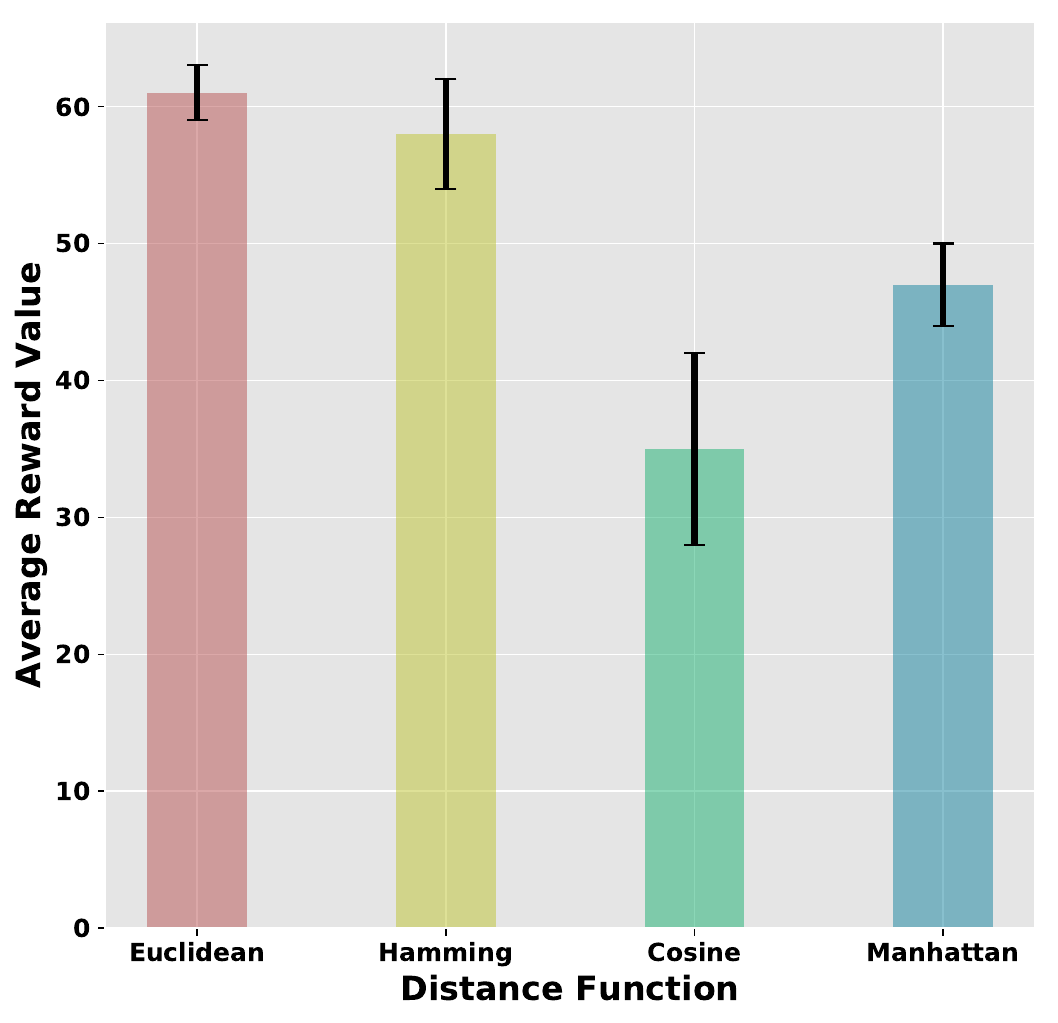}
     \subcaption{}\label{fig:DST}
   \end{minipage}
   \begin{minipage}[]{1\linewidth}
     \centering
     \includegraphics[scale=0.35]{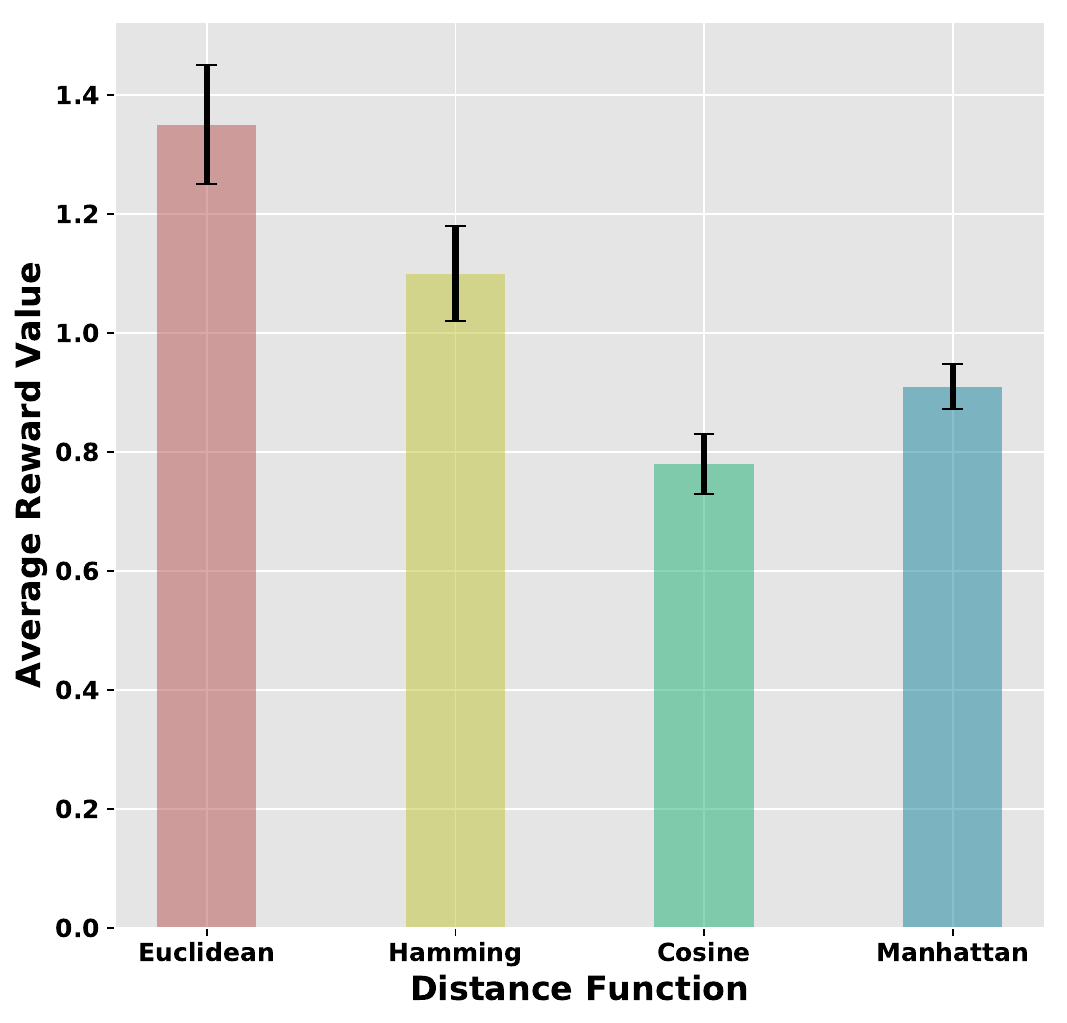}
     \subcaption{}\label{fig:RG}
   \end{minipage}
   \caption{Empirical analysis results for determining the distance function to be used in the RPM algorithm for each experimental environment. (a) Results for the SAR environment. (b) Results for the DST environment. (c) Results for the RG environment. The error bars represent the standard deviation.}
   \label{fig:distance_func}
\end{figure*}

Figure \ref{fig:distance_func} depicts the comparison results for different distance function and for each experimental environment.

\setcounter{subfigure}{0}

Based on the results it can be observed that the Euclidean distance function achieved the best overall results among the other comparative functions. A justification to this finding can be the representation of the user's preferences in our methodology as an Euclidean space of two dimensions ($w^1,w^2$) that are linearly separable. The RPB algorithm will utilize the Euclidean distance function during comparison to other algorithms.

\subsection{Performance Evaluation in Stationary Environments}

\begin{figure}
   \begin{minipage}[]{0.5\linewidth}
     \centering
     \includegraphics[width=9cm,height=7cm]{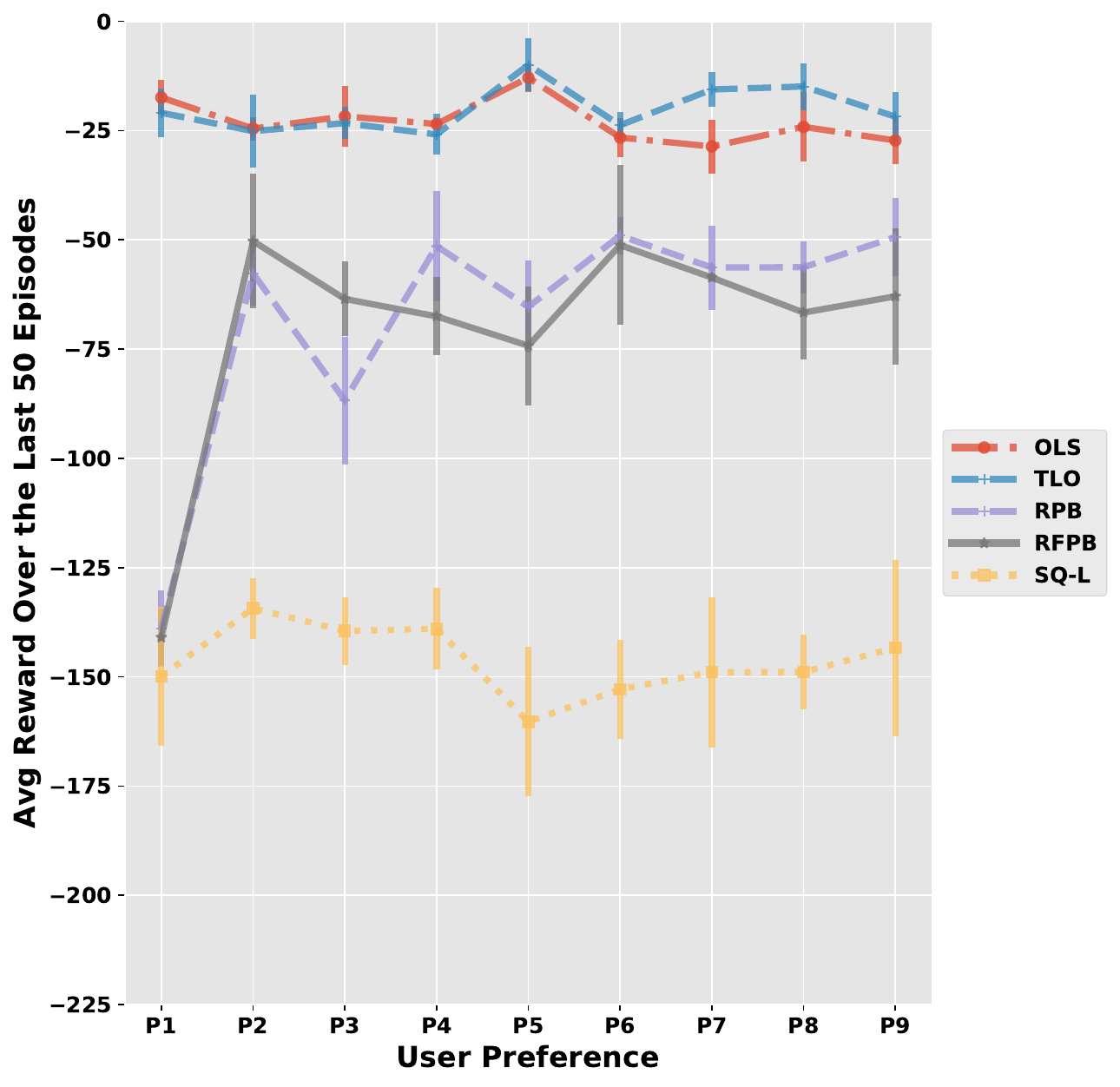}
     \subcaption{}\label{fig:a}
   \end{minipage}
   \begin{minipage}[]{0.5\linewidth}
     \centering
     \includegraphics[width=9cm,height=7cm]{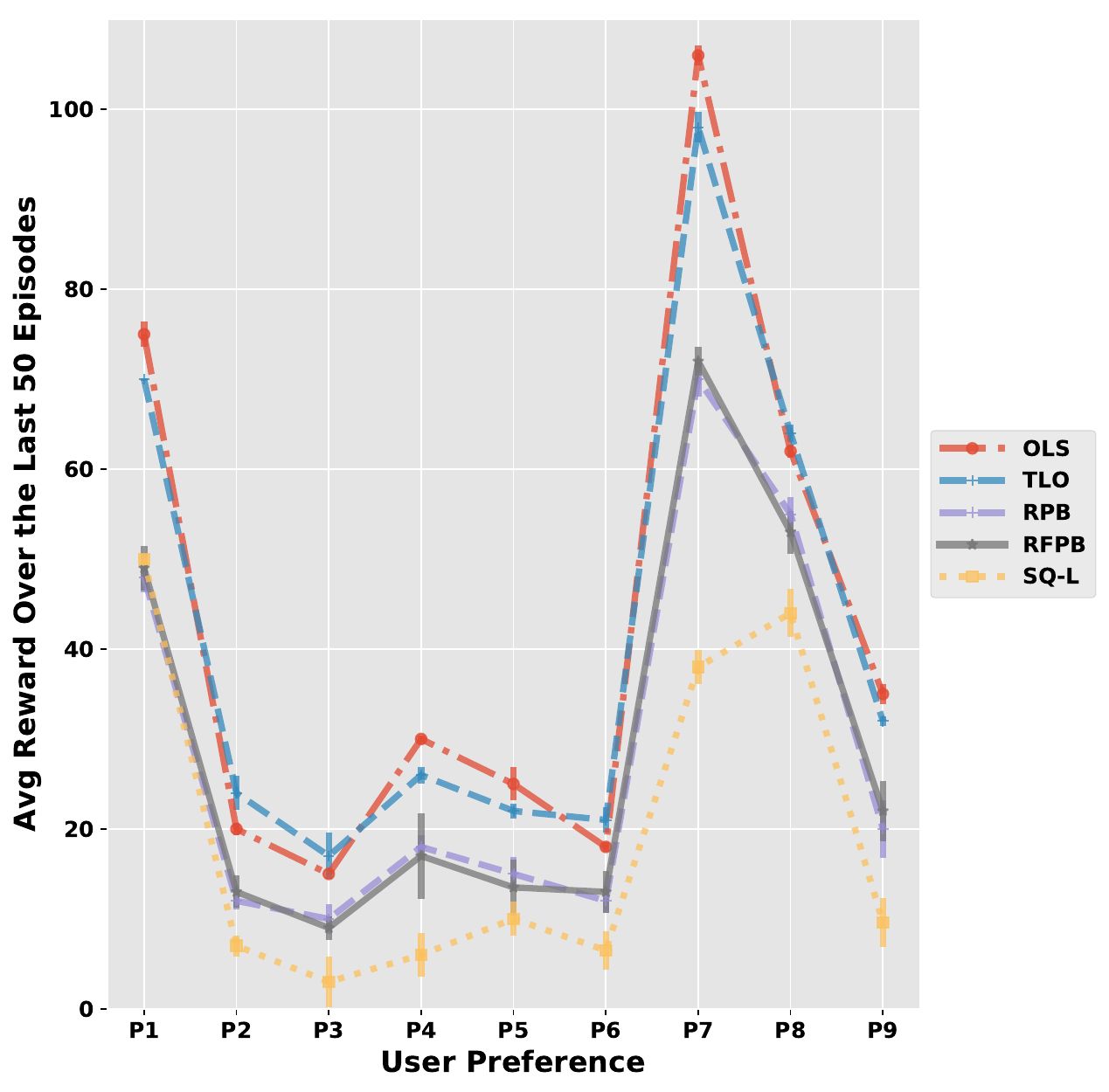}
     \subcaption{}\label{fig:b}
   \end{minipage}
   \begin{minipage}[]{0.5\linewidth}
     \centering
     \includegraphics[width=9cm,height=7cm]{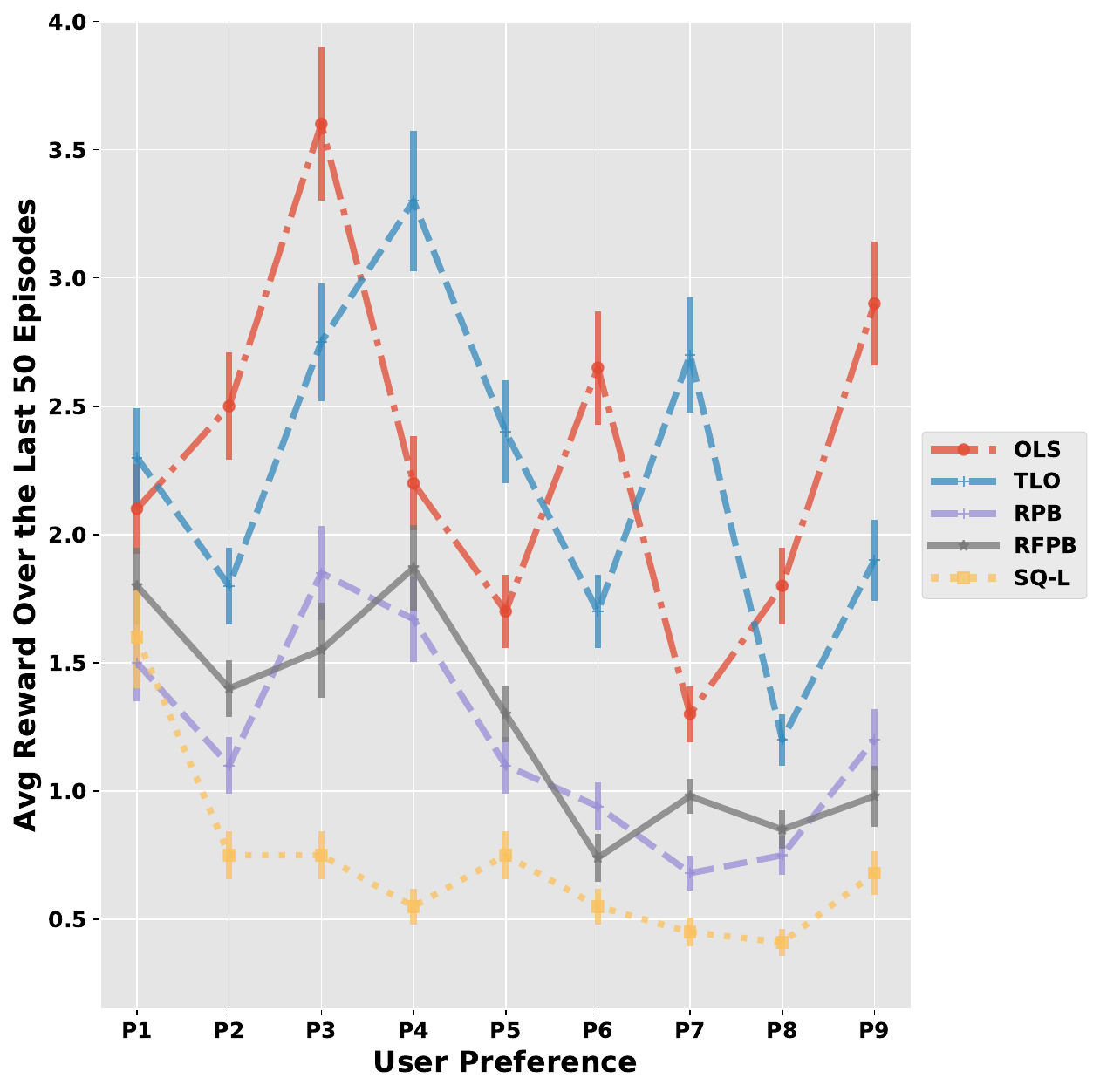}
     \subcaption{}\label{fig:c}
   \end{minipage}  
   \caption{Average reward value and standard deviation over $30$ runs achieved by the four algorithms for each preference in the stationary environments. (a) the  SAR environment. (b) the DST environment. (c) the RG environment.}
   \label{fig:Stationary_R_Result}
\end{figure}

We are going to present and discuss the results for each experimental environment for each of the two evaluation metrics utilized. For the average reward value over the last $50$ episodes metrics, Figure \ref{fig:Stationary_R_Result} shows a line plot that visualizes the average reward value and its standard deviation over $30$ runs for each experimental environment and for each comparison algorithm.

\setcounter{subfigure}{0}

While the results of the three environments differ in the magnitude of the average reward values due to changes in reward functions, they reflect common implications. It can be noticed that the OLS and TLO achieved the highest results in the average reward value across the three experimental environments, as the two algorithms evolved their CCS during the their offline training phases, therefore they were able to respond with an optimal policy after each preference change. While the RPB and RFPB algorithms started with an empty CCS as they work in an online manner, they achieved comparable results to the OLS and TLO algorithms and significantly (t-test: p-value  $< 0.05$) outperformed the SQ-L algorithm configured to respond to each preference change with a randomly initialized policy. The RPB algorithm benefits from the policy bootstrapping mechanism which enables it to build on previous experiences for each preference region given the threshold value ($\varphi$). The advantage of the RPB algorithm is its simplicity, while the RFPB algorithm removes requirement for the preference significance threshold parameter tuning.  

\begin{figure}
   \begin{minipage}[]{0.5\linewidth}
     \centering
     \includegraphics[width=9cm,height=7cm]{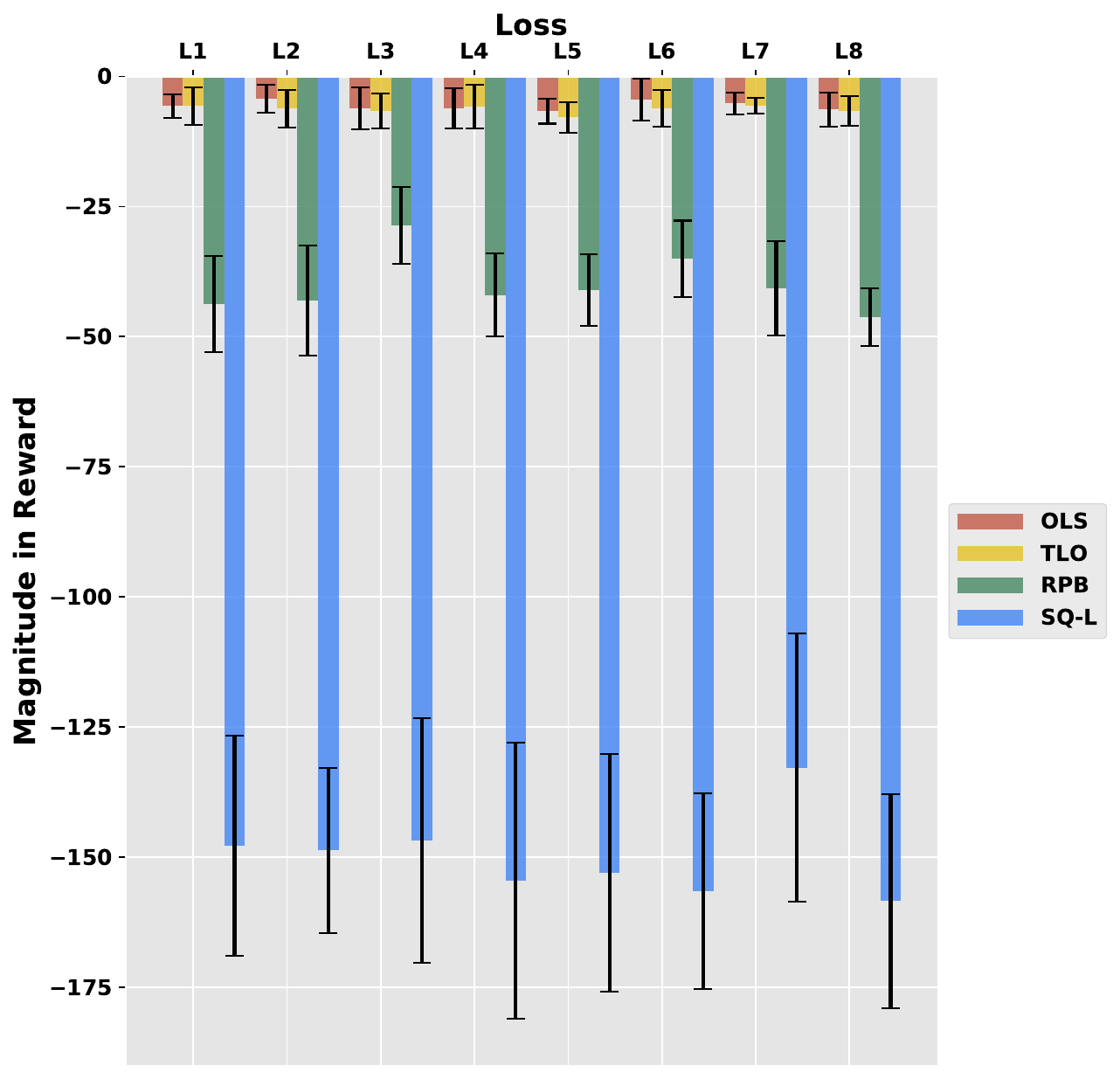}
     \subcaption{}\label{fig:a}
   \end{minipage}
   \begin{minipage}[]{0.5\linewidth}
     \centering
     \includegraphics[width=9cm,height=7cm]{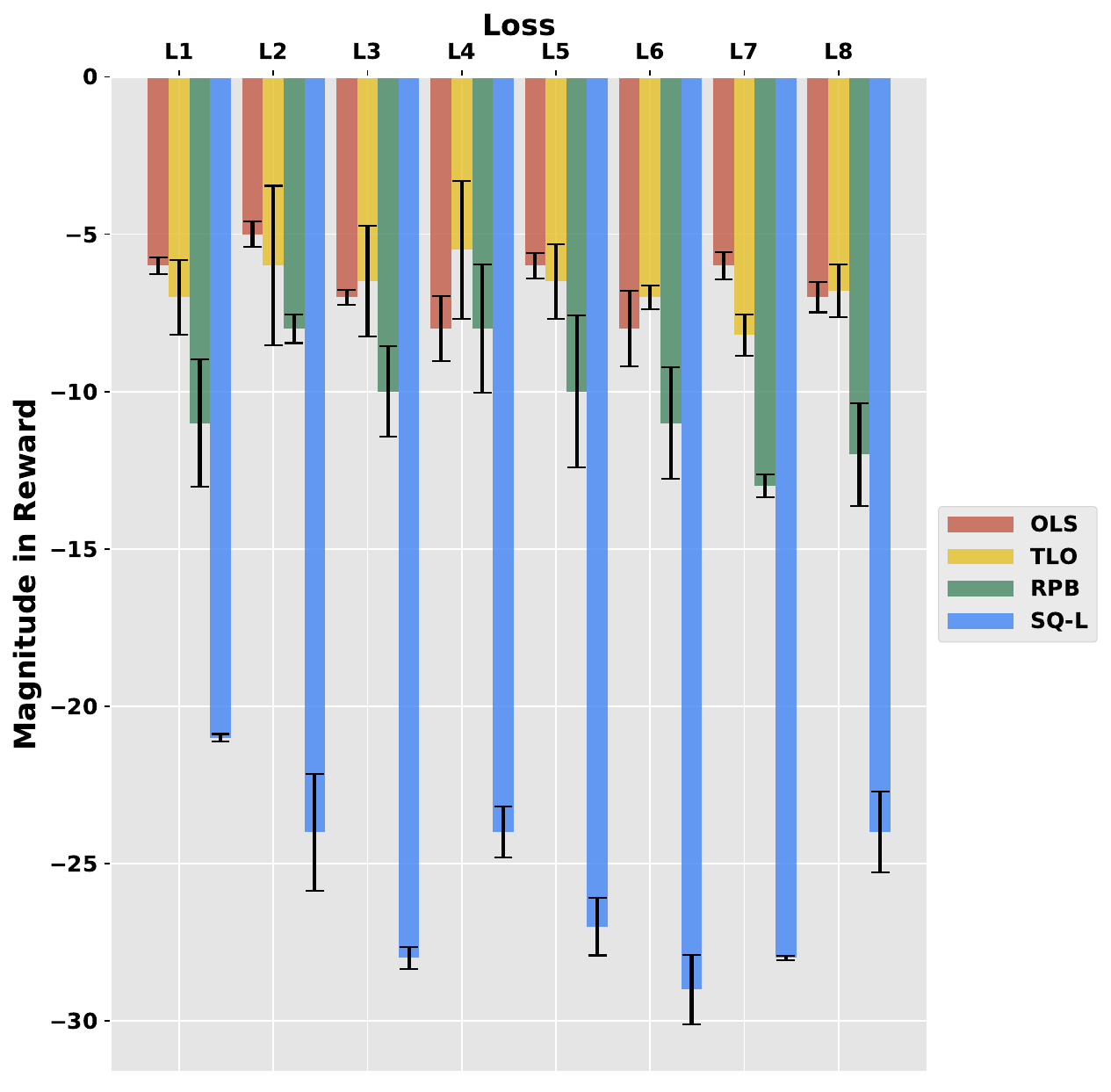}
     \subcaption{}\label{fig:b}
   \end{minipage}
   \begin{minipage}[]{0.5\linewidth}
     \centering
     \includegraphics[width=9cm,height=7cm]{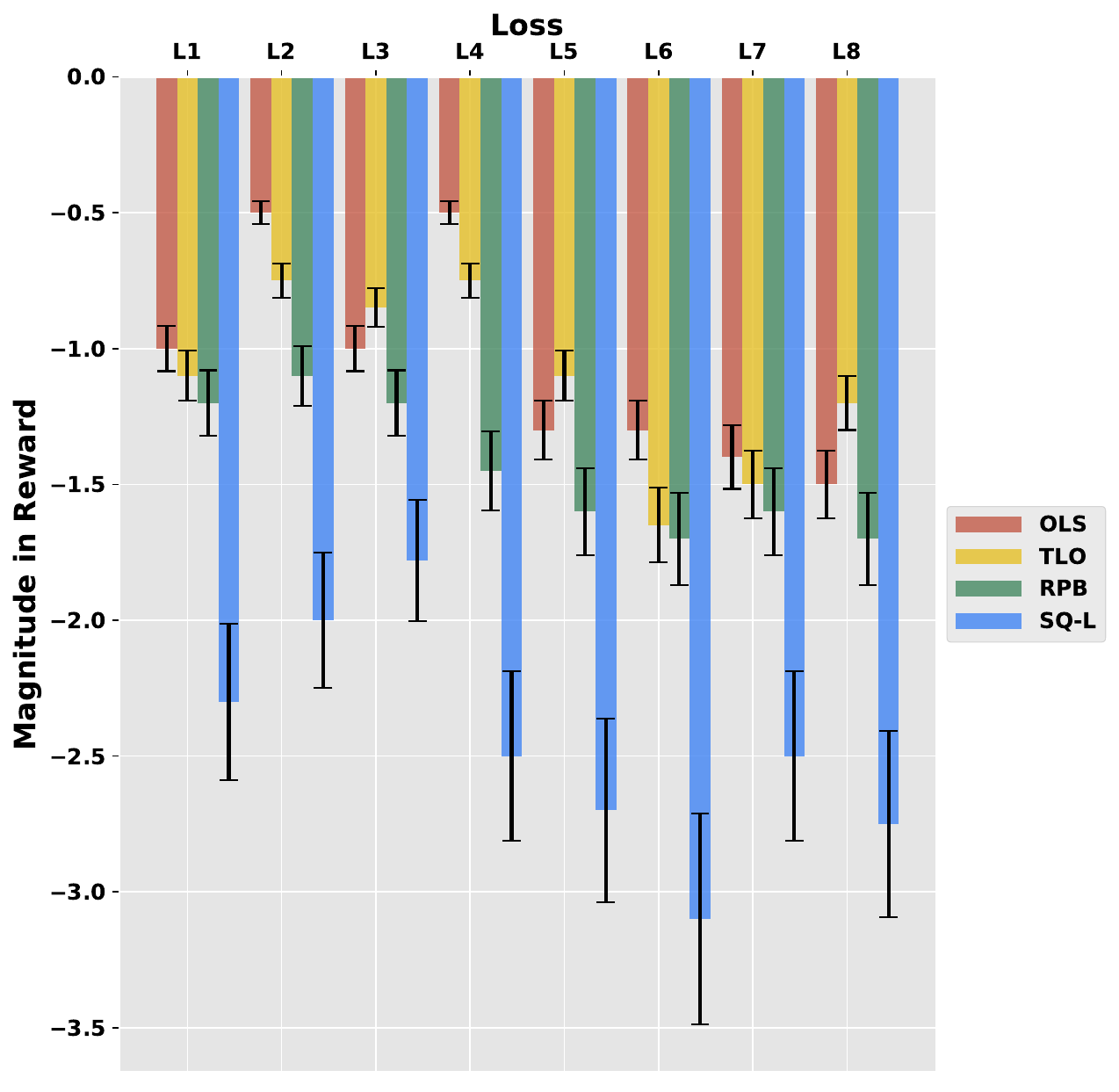}
     \subcaption{}\label{fig:c}
   \end{minipage}  
   \caption{Average reward loss after preference change with standard deviation over $30$ runs achieved by the four algorithms in the stationary environments. (a) the  SAR environment. (b) the DST environment. (c) the RG environment.}
   \label{fig:Stationary_loss_Result}
\end{figure}

For the average reward loss metric, Figure \ref{fig:Stationary_loss_Result} depicts the results of $30$ runs for each experimental environment and for each algorithm. As the RFPB achieved similar results on the average reward metric to the RPB, we only report the RPB results on the loss metric due to space constraints. Similarly, the average reward loss after each preference change emphasizes the findings from the average reward value metric as both OLS and TLO algorithms benefited from their evolved CCS during the offline training phase to minimize the reward loss after preference changes. While our RPB algorithm achieved comparable results in comparison to the OLS and TLO algorithms and significantly outperformed the SQ-L algorithm working with random initialization strategy after each preference change.  

\setcounter{subfigure}{0}

The results of the two metrics indicate that our RPB algorithm can achieve comparable results to the OLS and TLO algorithms that require an offline training phase to evolve their CCS, while our algorithm works in an online manner. Also, the RPB achieved a similar performance level compared to the RFPB algorithm, while it adopts a simpler preference decomposition technique. Moreover, the RPM algorithm outperforms the SQ-L baseline algorithms representing the single policy MORL approach that responds to each new preference by randomly initialized policy. While the results show that the current MORL multiple policy algorithms can deal with stationary environments through offline exhaustive search techniques, these environments are quite rare in practical problems. In the next experiment, we are going to assess the performance of the comparison algorithms under non-stationary environments to contrast the advantage of proposed RPB algorithm.

\subsection{Performance Evaluation in Non-stationary Environments}

\begin{figure}
   \begin{minipage}[]{0.5\linewidth}
     \centering
     \includegraphics[width=9cm,height=7cm]{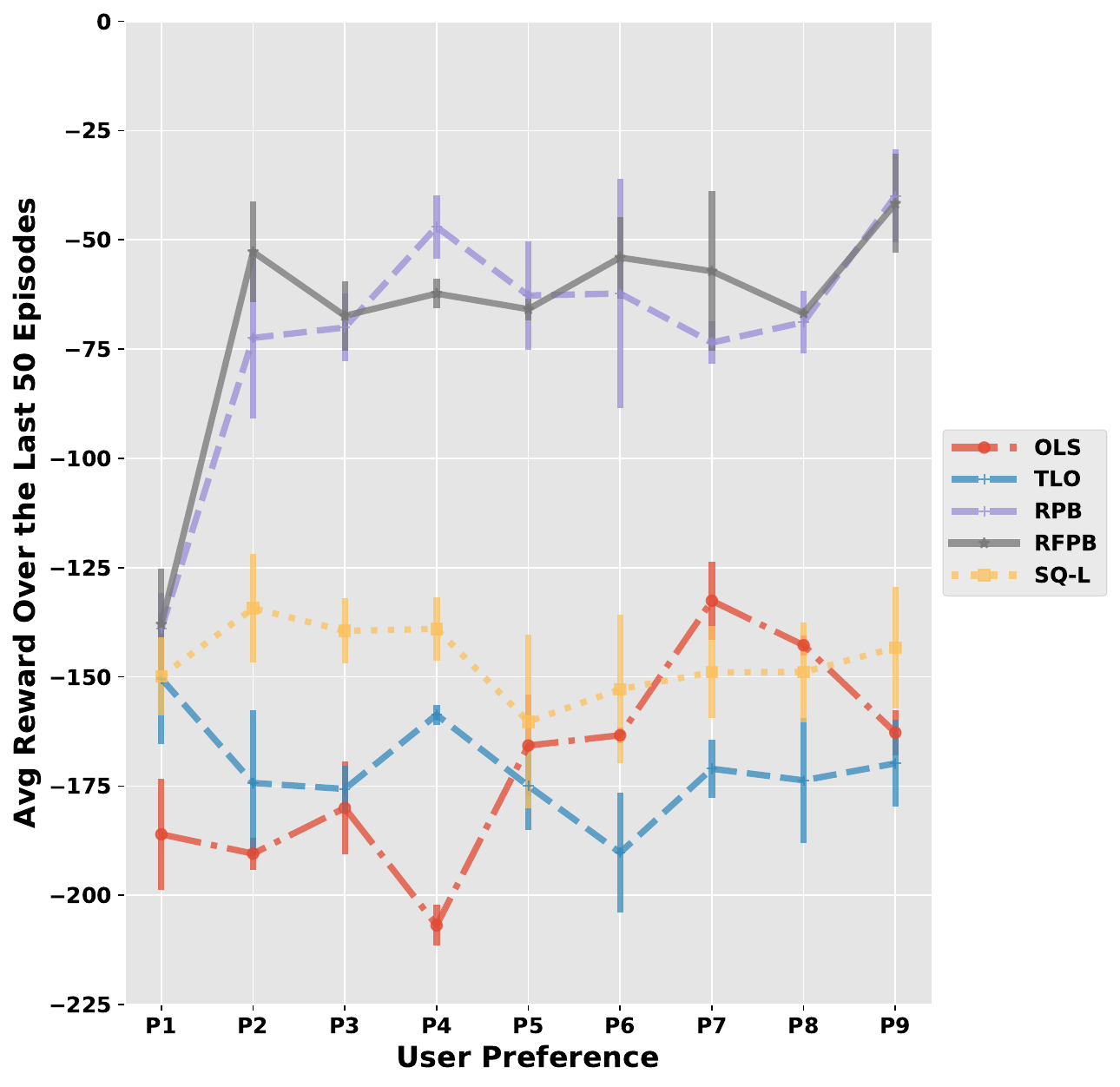}
     \subcaption{}\label{fig:a}
   \end{minipage}
   \begin{minipage}[]{0.5\linewidth}
     \centering
     \includegraphics[width=9cm,height=7cm]{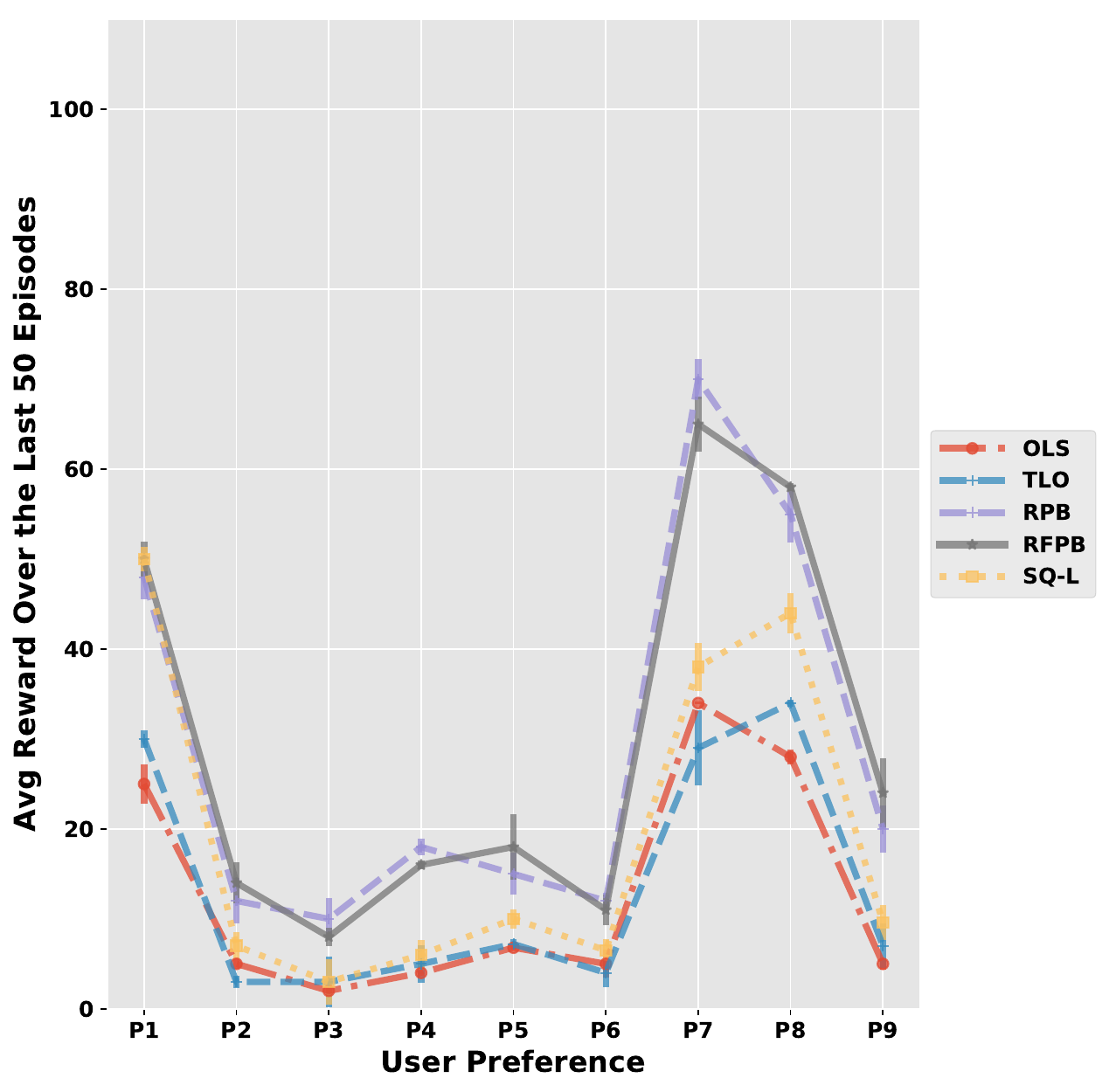}
     \subcaption{}\label{fig:b}
   \end{minipage}
   \begin{minipage}[]{0.5\linewidth}
     \centering
     \includegraphics[width=9cm,height=7cm]{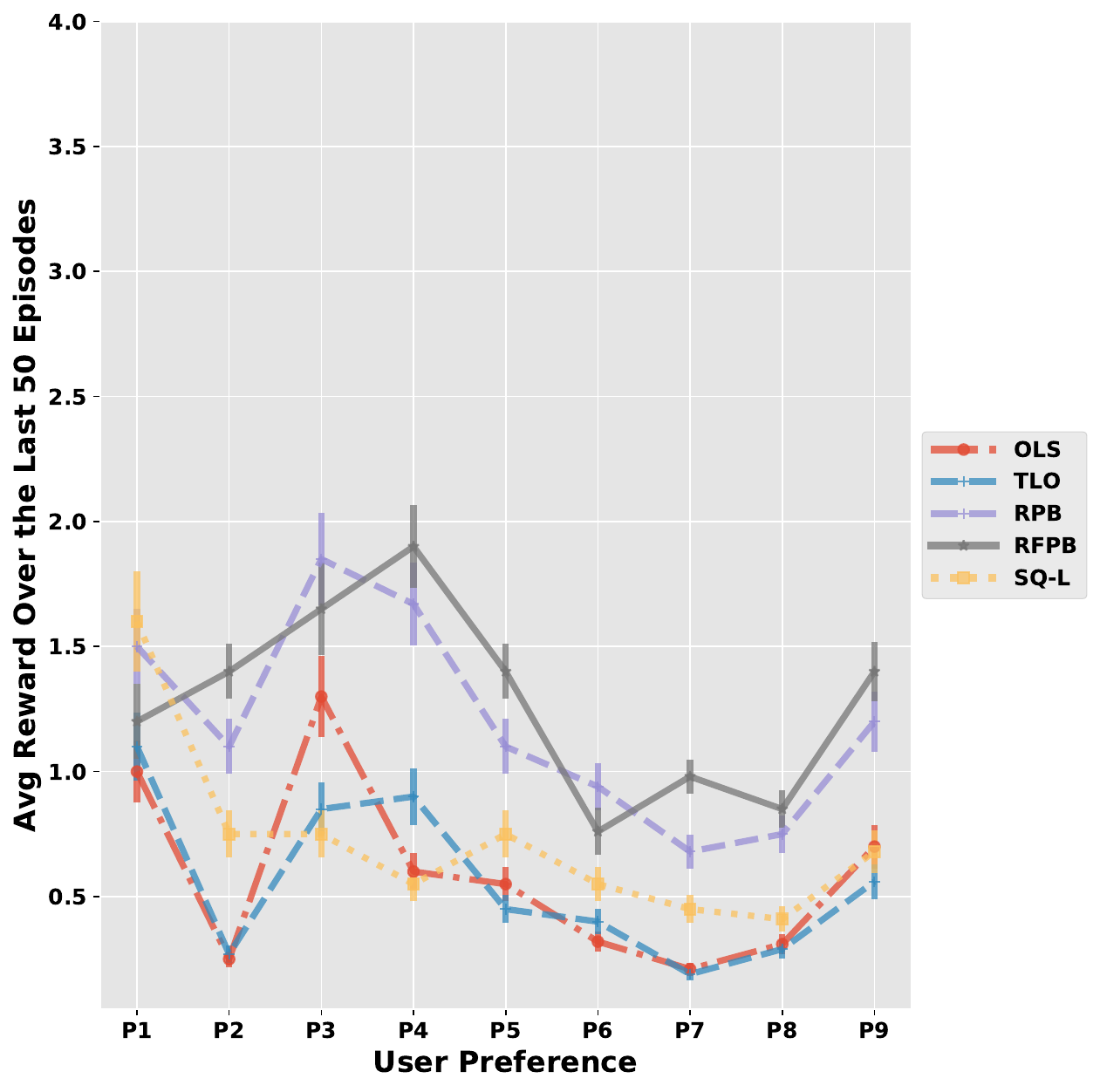}
     \subcaption{}\label{fig:c}
   \end{minipage}  
   \caption{Average reward value and standard deviation over $30$ runs achieved by the four algorithms for each preference in the non-stationary environments. (a) the  SAR environment. (b) the DST environment. (c) the RG environment.}
   \label{fig:Non-stationary_R_Result}
\end{figure}

For the average reward value over the last $50$ episodes metric, Figure \ref{fig:Non-stationary_R_Result} presents the comparison results for each experimental environment. Despite the variation in reward magnitude across environments as a result of different reward functions for each environment, the results show that the RPB algorithm performed comparably to the RFPB algorithm, while it significantly outperformed (t-test: p-value  $< 0.05$) the other comparative algorithms across all experimental environments. Mainly, this is due to its ability to evolve the CCS in an online manner, which enabled it to better cope with changes in the environment's dynamics in comparison to the OLS and TLO algorithms which failed to adapt due to their outdated CCS evolved in an offline manner. Also it can noticed that the SQ-L baseline, representing the single policy MORL approach, failed to adapt adequately as it can not accumulate knowledge from past experiences. Another remark in these results is that at some situations the OLS and TLO algorithms performed worse than the SQ-L baseline as their CCSs were greedy optimized for an outdated environment's setup, consequently, this experience was misleading during the new setup after changes in the environment's dynamics.     

\setcounter{subfigure}{0}

\begin{figure}
   \begin{minipage}[]{0.5\linewidth}
     \centering
     \includegraphics[width=9cm,height=7cm]{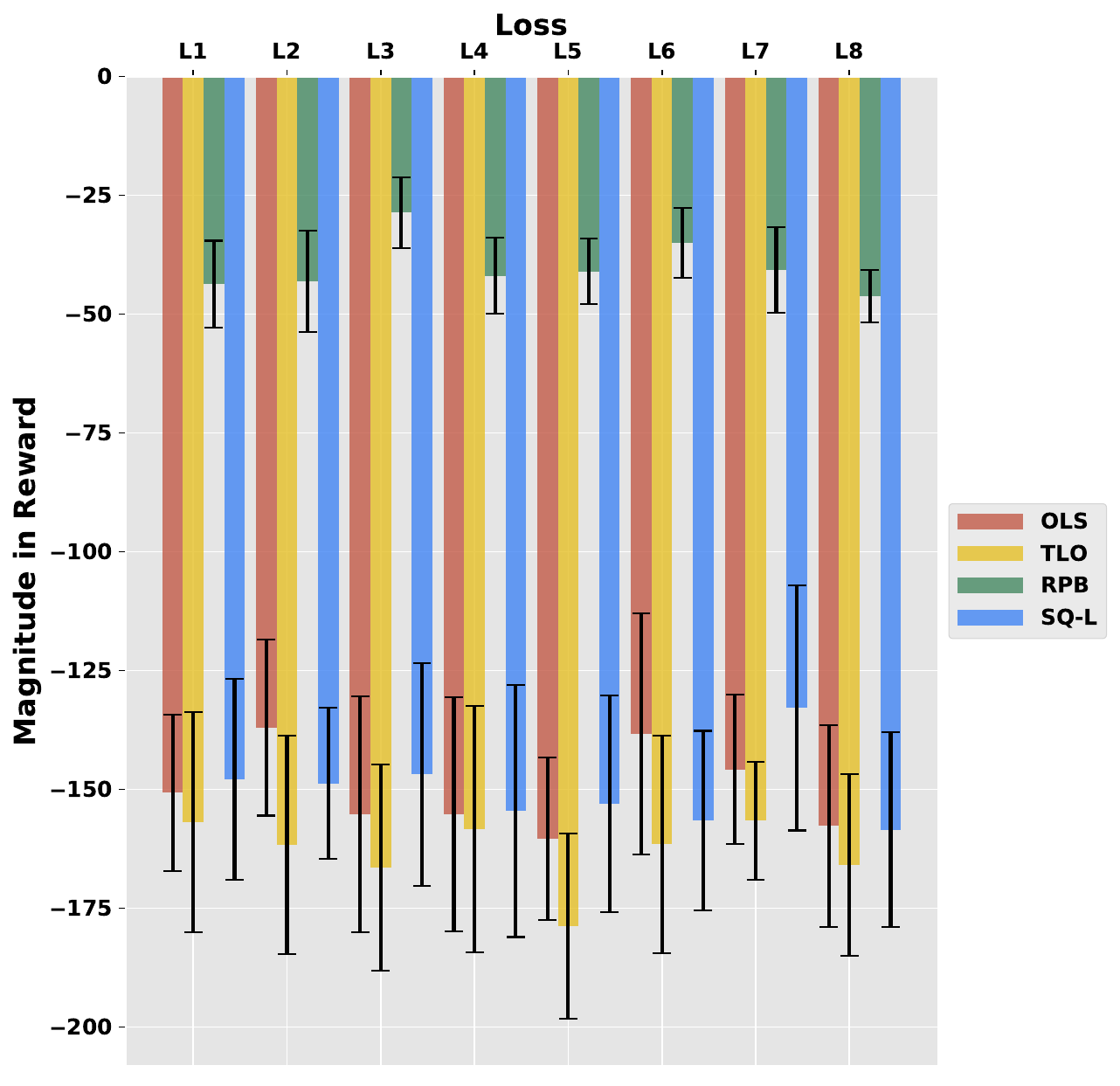}
     \subcaption{}\label{fig:a}
   \end{minipage}
   \begin{minipage}[]{0.5\linewidth}
     \centering
     \includegraphics[width=9cm,height=7cm]{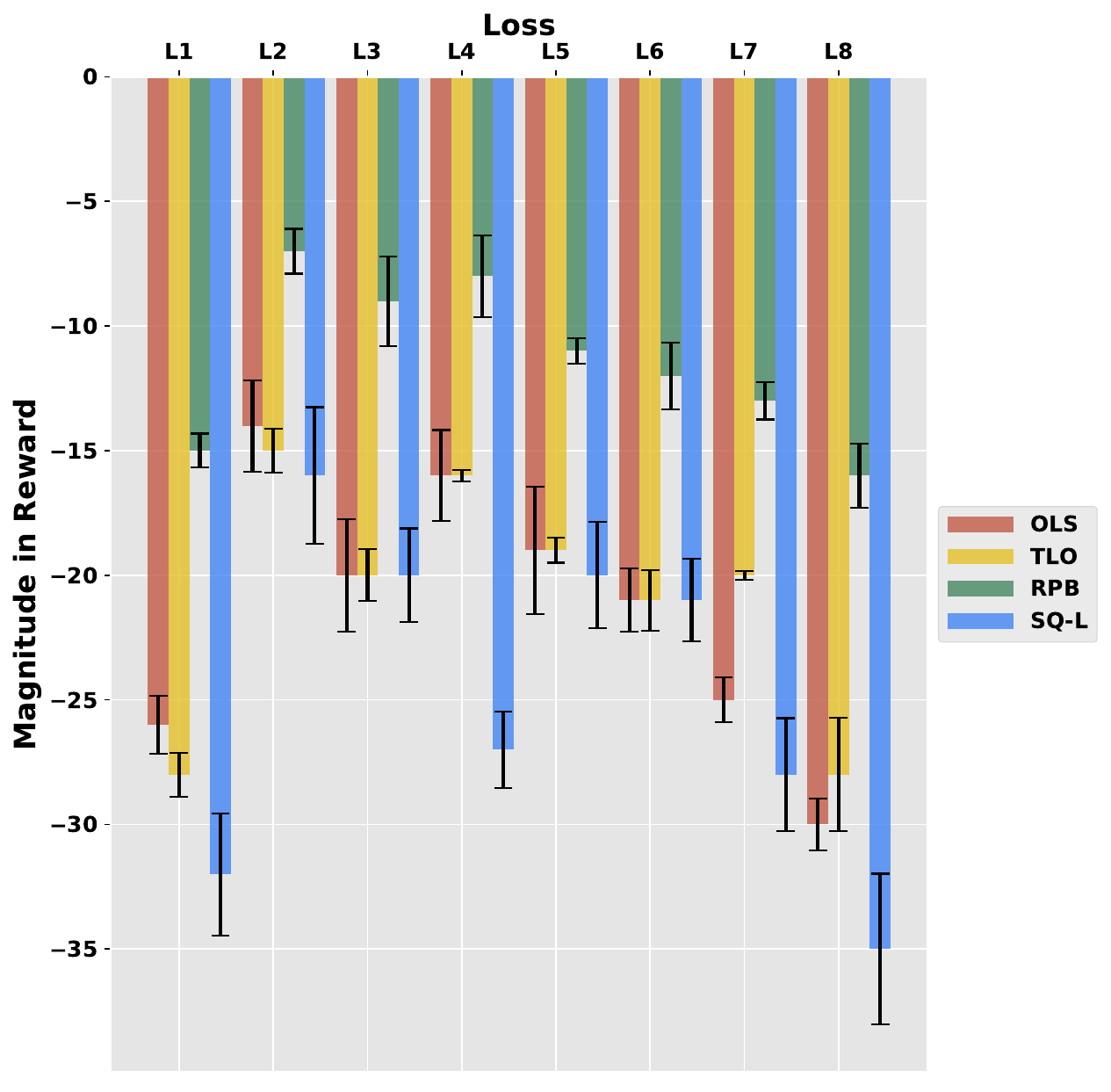}
     \subcaption{}\label{fig:b}
   \end{minipage}
   \begin{minipage}[]{0.5\linewidth}
     \centering
     \includegraphics[width=9cm,height=7cm]{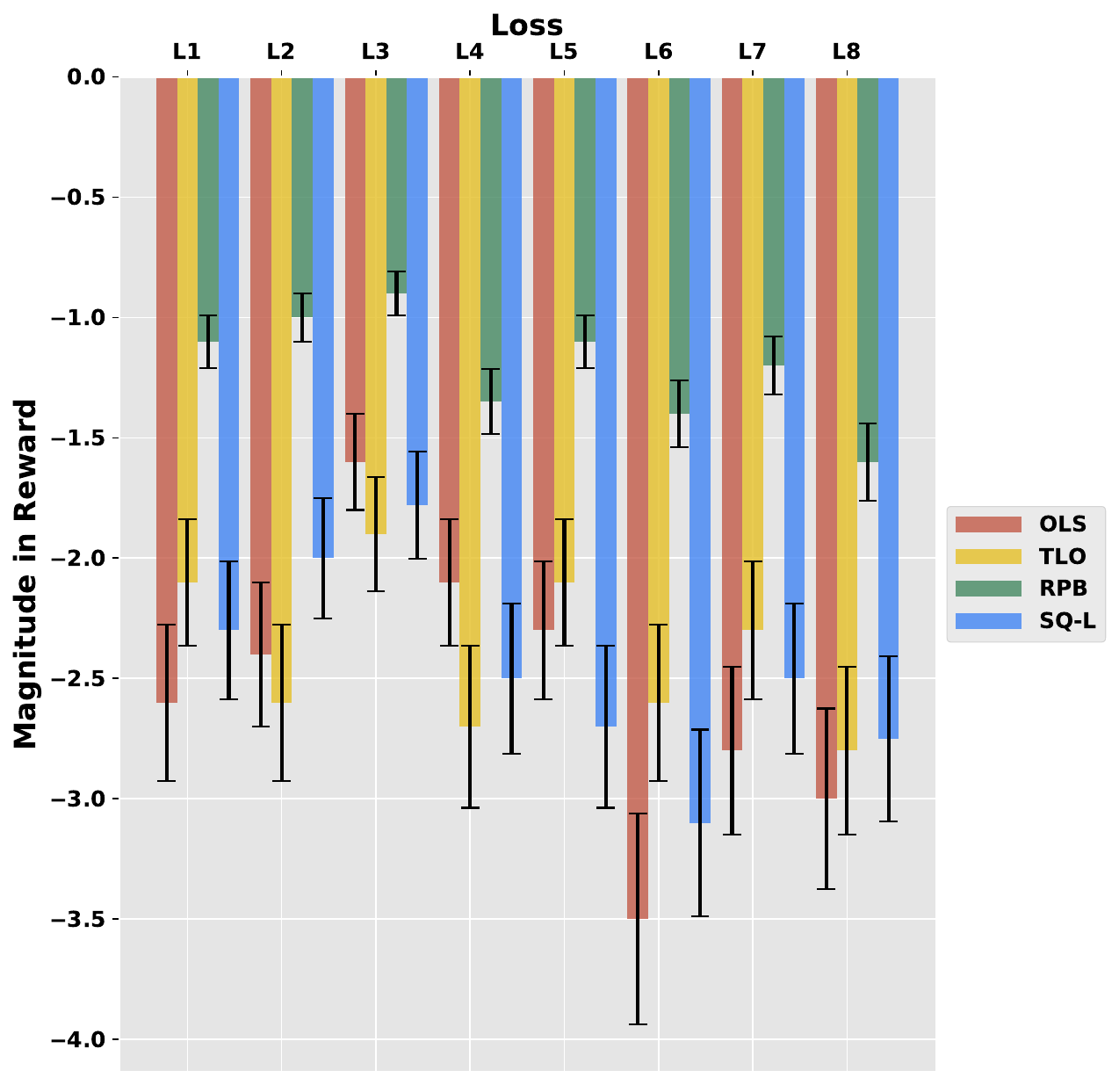}
     \subcaption{}\label{fig:c}
   \end{minipage}  
   \caption{Average reward loss after preference change with standard deviation over $30$ runs achieved by the four algorithms in the non-stationary environments. (a) the  SAR environment. (b) the DST environment. (c) the RG environment.}
   \label{fig:Nonstationary_loss_Result}
\end{figure}

The results of the average reward loss after preference change metric are visualized in Figure \ref{fig:Nonstationary_loss_Result}. These results show a similar message to the average reward value results as our RPB algorithm significantly outperformed (t-test: p-value  $< 0.05$) the other comparative algorithms in minimizing the reward loss after preference changes. While the two state-of-the-art multiple policy MORL algorithms (OLS and TLO) performed insignificantly to the SQ-L baseline due to the limitation to adapt with non-stationary dynamics in an online manner.   

\setcounter{subfigure}{0}

Based on the recent results, our RPB algorithm behaved in a more robust and adaptive manner in the non-stationary environments comparing to other comparative algorithms. Reasons behind this finding can be summarized as follows. First, the RPB algorithm targets generic steppingstone policies instead of specialized policies to specific environment setups in the case of the other algorithms, which makes its evolved coverage set more robust to changes in the environment dynamics. Second, the ability of the RPB algorithm to continuously evaluating and enhancing the policy set which cannot be achieved in the comparison algorithms that depends on offline and comprehensive policy search techniques.

\section{CONCLUSION AND FUTURE WORK} \label{sec:Conclusion}  

This paper proposed a novel algorithm that can robustly address multi-objective sequential decision-making problems in non-stationary environments.
This is achieved by evolving a robust steppingstone policy coverage set in an online
manner, then utilizing this coverage set to bootstrap specialized policies in response to changes in the user's preference or in the environment dynamics. We compared our proposed algorithm with state-of-the-art multi-objective reinforcement learning algorithms over three different multi-objective environments using both stationary and non-stationary dynamics.

We experimentally analyzed the different design decisions for the proposed algorithm in order to transparently describe the configuration setup and to indicate the possible configurable parts that can tailored for different application scenarios.
 
In order to contrast the performance of our proposed algorithm, we compared it with two state-of-the-art MORL algorithms and one baseline algorithm that is based on the well known Q-learning algorithm. We evaluated the comparison algorithms under both stationary and non-stationary environmental dynamics. During the stationary environments, the performance of the proposed algorithm was comparable to the performance of the other algorithms. While in the non-stationary environments,
the proposed algorithm significantly outperformed the other algorithms in terms of the average reward value achieved as it adapted better under changes in the environment dynamics.

The future extension for this work can be summarized into three main points. First, we are going to explore adaptive strategies for automatic preference exploration instead of the random strategy currently adopted. Unsupervised learning-based generative models can be for this purpose. Candidate strategies can include adversarial generative networks \citep{GANs} and intrinsically motivated exploration \citep{merrick2009motivated}. Second, we will investigate the impact of utilizing non-linear scalarization functions such as Chebyshev function, on the performance of the RPB algorithm. Finally, we are going to enhance the generalization ability of the RPB algorithm through adding a generic skill learning module, therefore, the generated skill set can be used to speed up the learning of specialized policies in different environments. We believe that equipping our proposed algorithm with the ability to learn generic skills will facilitate the evolution of better policies and the transfer of learned knowledge across different environments. 

\bibliographystyle{SageH}
\bibliography{RPB_Alg_Manuscript_Sherif.bib}


\end{document}